\documentclass{article}

\usepackage[eandd, preprint]{neurips_2026}

\usepackage[utf8]{inputenc}
\usepackage[T1]{fontenc}
\usepackage[colorlinks=true,
            linkcolor=linkskyblue,
            citecolor=linkskyblue,
            urlcolor=linkskyblue,
            filecolor=linkskyblue,
            breaklinks=true]{hyperref}
\usepackage{url}
\usepackage{booktabs}
\usepackage{amsfonts}
\usepackage{amsmath}
\usepackage{nicefrac}
\usepackage{microtype}
\usepackage{xcolor}
\usepackage{graphicx}
\usepackage{subcaption}
\usepackage{multirow}
\usepackage{enumitem}
\usepackage{colortbl}
\usepackage{array}
\usepackage{pifont}
\usepackage{amssymb}
\usepackage[most]{tcolorbox}

\newtcolorbox{examplebox}[1][]{%
  colback=gray!8, colframe=gray!40,
  fonttitle=\bfseries\small, boxrule=0.4pt,
  arc=2pt, left=4pt, right=4pt, top=3pt, bottom=3pt,
  #1
}

\definecolor{linkskyblue}{RGB}{30,144,220}

\definecolor{hlgray}{RGB}{235,235,235}
\definecolor{groupblue}{RGB}{225,236,247}   
\definecolor{groupred}{RGB}{247,229,229}    
\definecolor{humanbg}{RGB}{226,244,232}     
\definecolor{clusterbg}{RGB}{238,238,238}   
\definecolor{closedc}{RGB}{70,130,180}
\definecolor{openc}{RGB}{205,92,92}
\definecolor{humanc}{RGB}{60,179,113}
\definecolor{reasoningc}{RGB}{204,102,0}    
\definecolor{groundingc}{RGB}{40,116,166}   

\newcommand{\propgroup}[1]{\rowcolor{groupblue}\multicolumn{#1}{l}{\hspace{-0.3em}\textit{\textbf{Proprietary Models}}}}
\newcommand{\opengroup}[1]{\rowcolor{groupred}\multicolumn{#1}{l}{\hspace{-0.3em}\textit{\textbf{Open-source Models}}}}
\newcommand{\humanrow}[1]{\rowcolor{humanbg}\multicolumn{#1}{l}{\hspace{-0.3em}\textit{\textbf{Reference}}}}

\newcommand{\taskname}{Grounded Personality Reasoning}
\newcommand{\taskabbr}{GPR}
\newcommand{\benchname}{MM-OCEAN}
\newcommand{\gapname}{Rating--Grounding Misalignment}
\newcommand{\gapinv}{Prejudice Gap}
\newcommand{\best}[1]{\textbf{#1}}
\newcommand{\second}[1]{\underline{#1}}

\newcommand{\projectbadge}[3]{%
  \href{#3}{\raisebox{-0.18em}{#1}\hspace{2pt}\textcolor{linkskyblue}{\underline{#2}}}%
}

\setcitestyle{numbers,square,comma,sort&compress}

\newcommand{\figpipelinescale}{1.00}

\newcommand{\figframeworkscale}{1.00}

\newcommand{\tabmcqscale}{1.00}
\newcommand{\tableaderboardscale}{1.00}
\newcommand{\tabgapscale}{1.00}
\newcommand{\tabpercatscale}{1.00}

\newcommand{\tabreasoningscale}{1.00}
\newcommand{\tabtonescale}{1.00}      
\newcommand{\taboffnscale}{1.00}
\newcommand{\tabttwoscale}{1.00}      
\newcommand{\tabdifficultyscale}{1.00}
\newcommand{\tabpertraitscale}{1.00}
\newcommand{\tabscalingscale}{1.00}

\title{Perception or Prejudice: Can MLLMs Go Beyond First Impressions of Personality?}

\author{%
  \centerline{%
    \begin{tabular}{c}
      \rule{0pt}{20pt} \\[-10pt]
      Caixin Kang$^{1,2}$ \quad Tianyu Yan$^{2,3}$ \quad Sitong Gong$^{2,3}$ \quad Mingfang Zhang$^{1}$ \quad Liangyang Ouyang$^{1,2}$ \\[2pt]
      \textbf{Ruicong Liu}$^{1,2}$ \quad \textbf{Bo Zheng}$^{2}$ \quad \textbf{Huchuan Lu}$^{3}$ \quad \textbf{Kaipeng Zhang}$^{2}$ \quad \textbf{Yoichi Sato}$^{1}$ \quad \textbf{Yifei Huang}$^{1,2}$ \\[6pt]
      \normalfont\small $^{1}$The University of Tokyo \quad $^{2}$Shanda AI Research Tokyo \quad $^{3}$Dalian University of Technology \\[4pt]
      \normalfont\small \texttt{\{cxkang, ysato\}@iis.u-tokyo.ac.jp} \quad 
      \texttt{hyf015@gmail.com}%
    \end{tabular}%
  }%
}

\begin{document}

\maketitle

\begin{center}
\vspace{-1.4em}
\projectbadge{\includegraphics[height=0.95em]{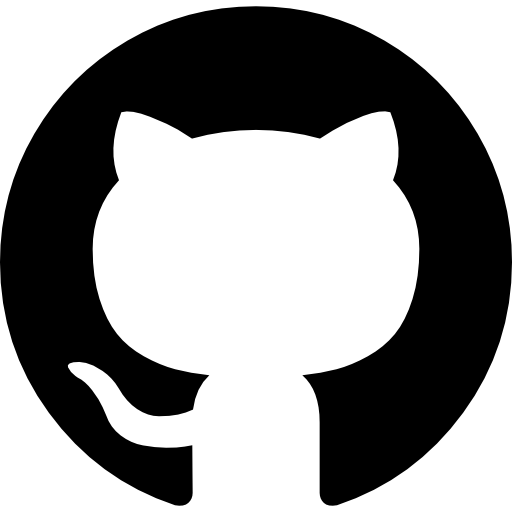}}{Code}{https://github.com/kkkcx/MM-OCEAN}
\quad
\projectbadge{\includegraphics[height=1.0em]{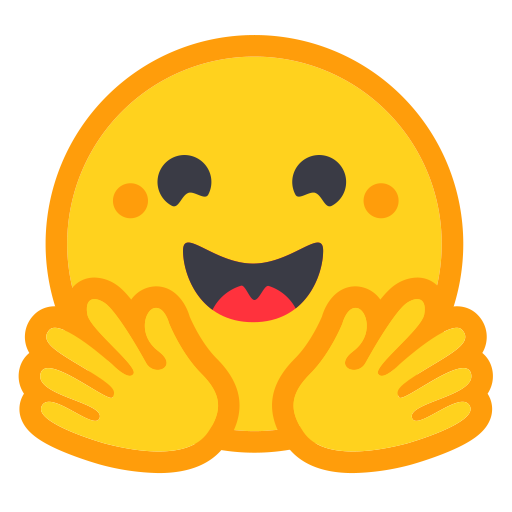}}{Dataset}{https://huggingface.co/datasets/anonymous-mm-ocean/MM-OCEAN}
\vspace{0.2em}
\end{center}

\begin{abstract}

Multimodal Large Language Models (MLLMs) are increasingly deployed in human-facing roles where \emph{personality perception} is critical, yet existing benchmarks evaluate this capability solely on numerical Big Five score prediction, leaving open whether models truly \emph{perceive} personality through behavioral understanding or merely \emph{prejudge} through superficial pattern matching. We address this gap with three contributions. \emph{(i) A new task:} we formalize \emph{\taskname{}} (\taskabbr{}), which requires MLLMs to anchor each Big Five rating in observable evidence through a chain of \emph{rating}, \emph{reasoning}, and \emph{grounding}. \emph{(ii) A new dataset:} we release \textbf{\benchname{}} (1{,}104 videos, 5{,}320 MCQs), produced by a multi-agent pipeline with human verification, with timestamped behavioral observations, evidence-grounded trait analyses, and seven categories of cue-grounding MCQs. \emph{(iii) Benchmark and analysis:} we design a three-tier evaluation (rating, reasoning, grounding) plus four sample-level failure-mode metrics: Prejudice Rate (PR), Confabulation Rate (CR), Integration-failure Rate (IR), and Holistic-grounding Rate (HR), and benchmark 27 MLLMs (13 closed, 14 open). The analysis uncovers a striking \emph{\gapinv{}}: across the field, $51\%$ of correct ratings are not grounded in retrieved cues, and the Holistic-Grounding Rate spans only $0$--$33.5\%$. 
These findings expose a disconnect between \emph{getting the right score} and \emph{reasoning for the right reason}, charting a roadmap for grounded social cognition in MLLMs.

\end{abstract}

\section{Introduction}
\label{sec:intro}

Multimodal Large Language Models (MLLMs) are rapidly entering high-stakes, human-centric applications: AI-powered interview screening~\cite{naim2016automated}, mental-health triage from facial and vocal cues~\cite{gratch2014distress}, social robots and companion digital humans that adapt to user traits~\cite{tang2025robot,cai2025towards}, and intelligent game NPCs that modulate behavior based on player affect~\cite{garavaglia2022moody5}. At the heart of all these systems lies a shared capability: \emph{personality perception}, the inference of stable psychological characteristics from observable behavior, with the Big Five (OCEAN) model~\cite{john2008paradigm} as the de facto target of inference.

But how well do current MLLMs actually understand the people they observe? Traditional benchmarks for apparent personality recognition (APR), such as ChaLearn First Impressions~\cite{ponce2016chalearn,escalante2020modeling}, frame the task as numerical regression on Big Five trait scores. This formulation cannot distinguish a model that ``gets it right'' from one that merely ``guesses right'': a model may achieve low prediction error by exploiting superficial correlations (e.g., smiling faces $\rightarrow$ high agreeableness) without genuinely understanding the supporting evidence, i.e., the right answer for the wrong reason.

This distinction between genuine \emph{perception} and superficial \emph{prejudice} carries practical stakes. 
Half a century of person-perception research shows that accurate trait inference rests on integrating specific behavioral micro-cues such as gaze and posture shifts, not on gestalt impressions~\cite{funder1995accuracy,ambady1992thin,liu2021generalizing}. By definition, a rating that cites no such cues is a prejudice, not a perception. Regulation has begun to enforce the same standard. The EU AI Act now classifies personality-based hiring and education systems as high-risk and mandates an explainable evidence trail for each deployed prediction~\cite{council2024regulation}. A personality judgment is trustworthy only if grounded in behavioral evidence. To formalize this requirement we introduce \emph{\taskname{} (\taskabbr{})}, which requires a model to (1)~\emph{perceive} fine-grained multimodal behavioral cues, (2)~\emph{reason} about how these cues map to personality traits via evidence-based analysis, and (3)~\emph{demonstrate} these abilities on structured multiple-choice probes that target specific sub-skills (e.g., microexpression localization, temporal-causal reasoning).

\begin{figure}[t]
\centering
\includegraphics[width=\figpipelinescale\linewidth]{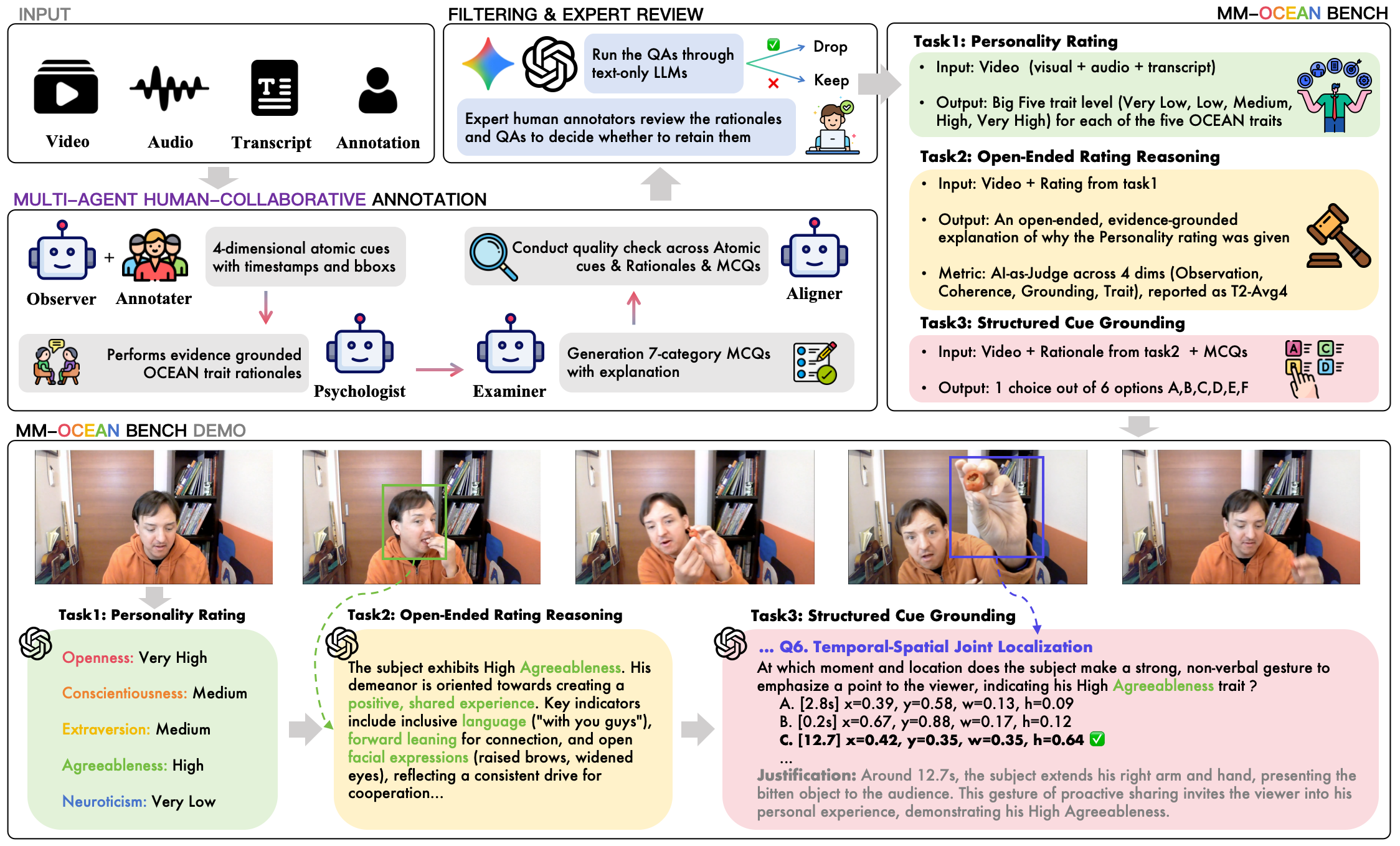}
\caption{\textbf{Overview of MM-OCEAN.} Multimodal inputs are processed by a multi-agent human-collaborative pipeline, filtered by text-only LLMs, and reviewed by experts to produce a benchmark supporting three tasks: ordinal Big Five rating (T1), open-ended evidence-grounded reasoning (T2), and structured cue-grounding Multiple-Choice Questions (MCQs) (T3).}
\label{fig:pipeline}
\vspace{-0.6cm}
\end{figure}

To evaluate \taskabbr{} we construct \benchname{}, comprising 1{,}104 videos and 5{,}320 cue-grounding MCQs built by a five-stage multi-agent human-collaborative annotation pipeline (Figure~\ref{fig:pipeline}). A three-tier evaluation framework probes the perception chain at increasing depth: \emph{ordinal personality rating} (Task~1), \emph{open-ended rating reasoning} (Task~2), and \emph{structured cue grounding} (Task~3; tested via targeted multiple-choice questions). Because aggregate task scores hide which step failed on a given sample, we add four sample-level failure-mode rates: \emph{Prejudice rate} (PR; right rating, wrong cues), \emph{Confabulation rate} (CR; plausible rationale, wrong cues), \emph{Integration-failure rate} (IR; right cues, wrong rating), and \emph{Holistic-Grounding rate} (HR; all three correct).

Benchmarking 27 representative MLLMs (13 proprietary, 14 open-source) reveals a striking \emph{\gapinv{}}: $51\%$ of all correct ratings come without grounded cue retrieval, and the Holistic-Grounding Rate spans only $0$--$33.5\%$. 
Moreover, recent \emph{reasoning-capable MLLMs} dominate the upper leaderboard, but the prejudice phenomenon is universal, even at the closed-source frontier, $\sim\!15\%$ of correct ratings remain ungrounded. 
Consequently, today's MLLMs often ``get the right score for the wrong reason,'' 
a gap our benchmark is designed to detect.
In summary, our contributions are as follows:
\begin{itemize}[leftmargin=*, itemsep=1pt]
\item \emph{Task.} We formalize \taskname{} (\taskabbr{}), distinguishing genuine \emph{perception} from \emph{prejudice} via a rating--reasoning--grounding chain.
\item \emph{Dataset.} We release \benchname{} (1{,}104 videos, 5{,}320 MCQs) with timestamped atomic observations, evidence-grounded trait analyses, and seven categories of cue-grounding MCQs, produced by an Observer--Psychologist--Examiner--Aligner pipeline with human verification.
\item \emph{Benchmark and analysis.} We design a three-tier evaluation framework (rating, reasoning, grounding) and four sample-level failure-mode metrics (PR/CR/IR/HR). Across 27 MLLMs we uncover the \gapinv{}, the discriminative power of HR, the prevalence of reasoning-capable variants among top performers, and two failure archetypes (\emph{confident raters} vs.\ \emph{cautious reasoners}).

\end{itemize}

\vspace{-0.2cm}
\section{Related Work}
\label{sec:related}

\vspace{-0.2cm}

\noindent\textbf{Psychological background: the Big Five model.}
The Big Five (OCEAN) model~\cite{mccrae1987validation,john2008paradigm} is the most empirically supported personality taxonomy in psychology, validated across languages and cultures and routinely used in clinical and social-science research~\cite{barrick1991bigfive}. Following ChaLearn First Impressions and most prior APR work, we adopt Big Five as the target of inference throughout \benchname{}.

\noindent\textbf{Apparent personality recognition.}
The ChaLearn Looking at People challenges~\cite{ponce2016chalearn,escalante2020modeling} established apparent personality recognition (APR), where models predict Big Five scores from short video clips via deep multimodal fusion~\cite{guclu2016deep}, from CNN aggregation~\cite{guclu2016deep} to Transformer architectures~\cite{saberi2026transformer}. All existing APR benchmarks remain pure regression with numerical labels, providing no mechanism to evaluate \emph{why} a particular score was assigned. \taskabbr{} reframes the task to require behaviorally grounded reasoning, not numerical outputs alone.

\noindent\textbf{Video understanding benchmarks for MLLMs.}
Recent benchmarks evaluate MLLMs' video understanding across temporal reasoning (TempCompass~\cite{liu2024tempcompass}, MVBench~\cite{li2024mvbench}), long-form comprehension (Video-MME~\cite{fu2024videomme}, EgoSchema~\cite{mangalam2023egoschema}), and multi-task assessment~\cite{fang2024mmbench}. While some touch on human-centric understanding through emotion recognition~\cite{poria2019meld} or action detection, none simultaneously target personality from video, require evidence-grounded reasoning, evaluate the reasoning chain itself, and supply fine-grained cue-grounding probes; \benchname{} fills these gaps along all four dimensions (Table~\ref{tab:related_bench}).

\noindent\textbf{Social cognition and theory of mind.}
ToM benchmarks (FANToM~\cite{kim2023fantom}, OpenToM~\cite{xu2024opentom}, Hi-ToM~\cite{wu2023hitom}) test reasoning about momentary mental states from text, and recent multimodal extensions probe higher-order social cognition such as deception in multi-party interactions~\cite{kang2025can} and multi-speaker attention ~\cite{ouyang2025multi,ouyang2026socialdirector}. 
Our work extends this line to personality perception, a higher-order social-cognitive task requiring multimodal integration over longer time spans and reasoning about stable trait dispositions; \taskabbr{} additionally requires reasoning to be \emph{grounded} in observable evidence.
\vspace{-0.2cm}

\begin{table}[t]
\caption{\textbf{Positioning of \benchname{} against related benchmarks.} 
\emph{Mod.}: V/A/T = video/audio/text; \emph{Format}: Reg/Cls/Open = regression/classification/open-ended.
}
\label{tab:related_bench}
\centering
\scriptsize
\setlength{\tabcolsep}{4pt}
\renewcommand{\arraystretch}{1.15}
\scalebox{0.85}{%
\begin{tabular}{llccccccc}
\toprule
\textbf{Benchmark} & \textbf{Domain} & \textbf{Mod.} & \textbf{Scale} & \textbf{Format} &
\textbf{Evid.} & \textbf{Reason.} & \textbf{Cue-Gnd.} & \textbf{Human} \\
& & & & & \textbf{grounded} & \textbf{eval} & \textbf{probes} & \textbf{val.} \\
\midrule
\rowcolor{clusterbg}\multicolumn{9}{l}{\hspace{-0.3em}\textit{\textbf{Apparent personality recognition (video)}}}\\
ChaLearn FI v1/v2~\cite{ponce2016chalearn,escalante2020modeling}        & Big Five              & V+A   & $\sim$10k     & Reg      & \textcolor{openc}{\ding{55}} & \textcolor{openc}{\ding{55}} & \textcolor{openc}{\ding{55}} & partial \\
UDIVA~\cite{palmero2021udiva}                    & Big Five (dyadic)     & V+A   & 188 sessions  & Reg      & \textcolor{openc}{\ding{55}} & \textcolor{openc}{\ding{55}} & \textcolor{openc}{\ding{55}} & partial \\
\midrule
\rowcolor{clusterbg}\multicolumn{9}{l}{\hspace{-0.3em}\textit{\textbf{General-purpose video benchmarks for MLLMs}}}\\
MVBench~\cite{li2024mvbench}                  & 20 video tasks        & V     & 4k MCQ        & MCQ      & \textcolor{openc}{\ding{55}} & \textcolor{openc}{\ding{55}} & partial (by task) & partial \\
Video-MME~\cite{fu2024videomme}                & 30 subtasks           & V+A   & 2.7k MCQ      & MCQ      & \textcolor{openc}{\ding{55}} & \textcolor{openc}{\ding{55}} & partial (by domain) & \textcolor{humanc}{\checkmark} \\
TempCompass~\cite{liu2024tempcompass}              & Temporal video        & V     & 7k items      & Mixed    & \textcolor{openc}{\ding{55}} & \textcolor{openc}{\ding{55}} & partial (5 asp.) & \textcolor{humanc}{\checkmark} \\
Perception Test~\cite{patraucean2023perception}          & Perceptual video      & V+A   & 11.6k         & MCQ+Loc. & \textcolor{openc}{\ding{55}} & \textcolor{openc}{\ding{55}} & partial & \textcolor{humanc}{\checkmark} \\
\midrule
\rowcolor{clusterbg}\multicolumn{9}{l}{\hspace{-0.3em}\textit{\textbf{Social cognition / Theory of Mind (text-only)}}}\\
SocialIQA~\cite{sap2019socialiqa}                & Social commonsense    & T     & 38k           & MCQ      & \textcolor{openc}{\ding{55}} & \textcolor{openc}{\ding{55}} & \textcolor{openc}{\ding{55}} & \textcolor{humanc}{\checkmark} \\
FANToM~\cite{kim2023fantom}                   & ToM in dialogue       & T     & $\sim$10k     & Open+MCQ & \textcolor{openc}{\ding{55}} & partial & \textcolor{openc}{\ding{55}} & \textcolor{humanc}{\checkmark} \\
Hi-ToM~\cite{wu2023hitom}                   & Higher-order ToM      & T     & 600           & MCQ      & \textcolor{openc}{\ding{55}} & \textcolor{openc}{\ding{55}} & \textcolor{openc}{\ding{55}} & \textcolor{openc}{\ding{55}} \\
\midrule
\rowcolor{clusterbg}\multicolumn{9}{l}{\hspace{-0.3em}\textit{\textbf{Video emotion / affective understanding}}}\\
MELD~\cite{poria2019meld}                     & Conversational emotion& V+A+T & 13k utter.    & Cls      & \textcolor{openc}{\ding{55}} & \textcolor{openc}{\ding{55}} & \textcolor{openc}{\ding{55}} & \textcolor{humanc}{\checkmark} \\
EmoBench~\cite{sabour2024emobench}                 & Emotional intelligence& T     & 400           & MCQ      & \textcolor{openc}{\ding{55}} & partial & partial (2 asp.) & \textcolor{humanc}{\checkmark} \\
\midrule
\rowcolor{humanbg}\textbf{\benchname{} (ours)}
                                 & \textbf{GPR (Big Five + cue-grounding)}
                                                        & \textbf{V+A+T}
                                                                & \textbf{1{,}104 / 5{,}320}
                                                                                & \textbf{Multi}
                                                                                        & \textcolor{humanc}{\checkmark}
                                                                                                      & \textcolor{humanc}{\checkmark}
                                                                                                                    & \textcolor{humanc}{\checkmark\,(7 cat.)}
                                                                                                                                  & \textcolor{humanc}{\checkmark} \\
\bottomrule
\end{tabular}%
}
\vspace{-0.3cm}
\end{table}

\section{\benchname{}: Benchmark Construction}
\label{sec:benchmark}

\subsection{Task Definition: \taskname{}}
\label{sec:task_def}

\noindent\textbf{Input.}  A Grounded Personality Reasoning (\taskabbr{}) instance is a short video $V = (V_{\text{vis}}, V_{\text{aud}}, V_{\text{txt}})$ comprising a sequence of RGB frames $V_{\text{vis}} \in \mathbb{R}^{T\times H\times W\times 3}$, an audio waveform $V_{\text{aud}}$, and a speech transcription $V_{\text{txt}}$. We denote the set of Big Five traits by $\mathcal{T} = \{E, A, C, N, O\}$ and the ordinal personality scale by $\mathcal{L} = \{1,2,3,4,5\}$ (Very Low to Very High).

\noindent\textbf{Outputs across the three tasks.} A model $f_\theta$ must produce:
\begin{align}
\text{\small\textbf{T1} (Rating)}\quad
& \hat{y}_i \in \mathcal{L},\ \ \forall\, i \in \mathcal{T}, \label{eq:t1}\\[-2pt]
\text{\small\textbf{T2} (Reasoning)}\quad
& (\hat{\mathcal{O}}, \hat{\mathcal{R}}) = f_\theta(V), \quad
  \hat{\mathcal{O}} = \{o_k\}_{k=1}^{K},\ \
  \hat{\mathcal{R}} = \{r_i \mid i\!\in\!\mathcal{T}\}, \label{eq:t2}\\[-2pt]
\text{\small\textbf{T3} (Grounding)}\quad
& \hat{a}_q \in \{\texttt{A},\texttt{B},\texttt{C},\texttt{D},\texttt{E},\texttt{F}\}, \quad \forall\, q \in \mathcal{Q}, \label{eq:t3}
\end{align}
where each observation $o_k\!=\!(d_k, t^s_k, t^e_k, \text{desc}_k, b_k)$ records a perceptual dimension $d_k\!\in\!\{\text{Expression, Action, Audio, Background}\}$, start/end timestamps (in seconds), a free-text description $\text{desc}_k$, and a body-part tag $b_k$ (e.g., face, hand); each reasoning chain $r_i\!=\!(\ell_i, \mathcal{E}_i, \text{rat}_i)$ comprises the predicted trait level $\ell_i\!\in\!\mathcal{L}$, an evidence set $\mathcal{E}_i\!\subseteq\!\{1,\dots,K\}$ of observation indices (\emph{OBS-IDs}), and a free-text rationale $\text{rat}_i$; $\mathcal{Q}$ is the set of seven cue-grounding MCQs for $V$. The \emph{grounding constraint} $\mathcal{E}_i\!\subseteq\!\{1,\dots,K\}$ --- every trait judgment must cite at least one observed cue --- is what distinguishes \taskabbr{} from Apparent Personality Recognition (APR), which evaluates only $\hat{y}_i$.

\begin{table}[t]
\caption{\textbf{Seven cue-grounding MCQ categories generated by the Examiner.} Two clusters: \textcolor{reasoningc}{Reasoning} (semantic / causal inference) and \textcolor{groundingc}{Visual Grounding} (pixel- / time-level localization).}
\label{tab:mcq_taxonomy}
\centering
\footnotesize
\setlength{\tabcolsep}{5pt}
\renewcommand{\arraystretch}{0.95}
\scalebox{\tabmcqscale}{%
\begin{tabular}{@{}>{\raggedright\arraybackslash}p{3.3cm} >{\raggedright\arraybackslash}p{3.4cm} p{5.6cm}@{}}
\toprule
\textbf{Category} & \textbf{Cognitive target} & \textbf{Example} \\
\midrule
\rowcolor{clusterbg}\multicolumn{3}{@{}l}{\hspace{-0.3em}\textbf{\textcolor{reasoningc}{Reasoning cluster}}}\\
\textbf{Personality Attrib.}\,\textcolor{reasoningc}{\scriptsize(\textit{Pers})}
  & Behavior $\to$ trait mapping
  & \textit{``Which Big Five trait does the behavior at $11.6$--$14.8$s most strongly support?''} \\
\addlinespace[2pt]
\textbf{Counterfactual}\,\textcolor{reasoningc}{\scriptsize(\textit{Counter})}
  & Alternative-scenario inference
  & \textit{``If the behavior at $11.6$--$14.8$s were absent, which trait rating would change most?''} \\
\addlinespace[2pt]
\textbf{Temporal-Causal}\,\textcolor{reasoningc}{\scriptsize(\textit{TempC})}
  & Cause-effect across time
  & \textit{``Which causal chain best links the person's actions across the video?''} \\
\addlinespace[2pt]
\textbf{Mixed Emotion}\,\textcolor{reasoningc}{\scriptsize(\textit{Mixed})}
  & Complex affective state
  & \textit{``During $1.9$--$8.4$s, the person's emotional state is best characterized as $\ldots$''} \\
\midrule
\rowcolor{clusterbg}\multicolumn{3}{@{}l}{\hspace{-0.3em}\textbf{\textcolor{groundingc}{Visual Grounding cluster}}}\\
\textbf{Micro-expression}\,\textcolor{groundingc}{\scriptsize(\textit{Micro})}
  & Subtle facial signal detection
  & \textit{``When does a notable micro-expression change relevant to the rated trait occur?''} \\
\addlinespace[2pt]
\textbf{Spatial Loc.}\,\textcolor{groundingc}{\scriptsize(\textit{Spat})}
  & Body-part-level localization
  & \textit{``At ${\sim}12.6$s within bbox $(0.30, 0.66, 0.09, 0.09)$, what is the most prominent action?''} \\
\addlinespace[2pt]
\textbf{Temp-Spatial Jnt.}\,\textcolor{groundingc}{\scriptsize(\textit{TSJnt})}
  & Joint time$\times$space grounding
  & \textit{``When and where does the subject make a head-coordinated emphatic gesture?''} \\
\bottomrule
\end{tabular}%
}
\vspace{-0.3cm}
\end{table}

\begin{figure}[t]
\centering
\includegraphics[width=0.98\linewidth]{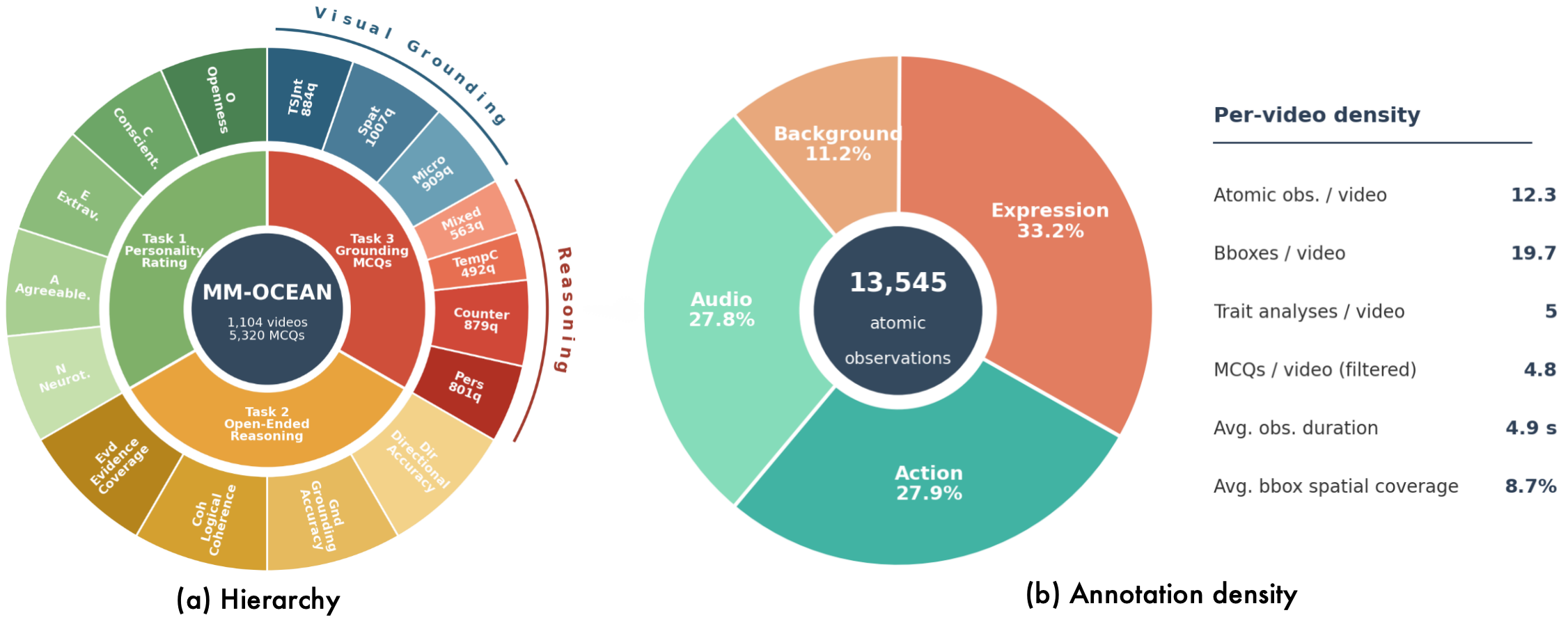}
\vspace{-0.2cm}
\caption{\textbf{\benchname{} overview.} (a)~Three-layer sunburst over benchmark scope, three evaluation tasks, and the seven cue-grounding categories. (b)~Atomic-observation density across the four perceptual channels; bounding-box geometry is attached to every Expression / Action observation.}
\label{fig:overview}
\vspace{-0.6cm}
\end{figure}

\vspace{-0.3cm}
\subsection{Multi-Agent Human-Collaborative Annotation Pipeline}
\label{sec:pipeline}
\vspace{-0.2cm}
\benchname{} is constructed through a five-stage pipeline that interleaves four LLM agents (\emph{Observer}, \emph{Psychologist}, \emph{Examiner}, and \emph{Aligner}) with two complementary human roles: 24 trained \emph{annotator-verifiers} (Stage~1) and a pool of \emph{expert reviewers} (Stage~5), as visualized in Figure~\ref{fig:pipeline}. The full annotation protocol, web-tool design, and inter-annotator agreement are detailed in Appendix~\ref{app:human}.

\noindent\textbf{Stage 1. Atomic-Cue Annotation (Observer + Human).}
\label{sec:human}
The \emph{Observer} agent receives the video and transcription and emits \emph{atomic behavioral observations}, i.e., the smallest indivisible behavioral events (e.g., a single eyebrow raise, a brief pause), each tagged with a unique OBS-ID, a perceptual dimension (Expression, Action, Audio, Background), preliminary timestamps, a factual description, and body-part labels. 
24 trained human annotators then review every drafted cue, labelling it \emph{correct}, \emph{incorrect}, or \emph{nonexistent} and pruning the latter two; for every retained Expression or Action observation, the annotator further refines its timestamps and tight bounding-box via a frame-accurate web tool we built. $78.2\%$ of Observer drafts are accepted, $14.6\%$ corrected, and $5.9\%$ deleted; pairwise verdict agreement on overlap pool is $77\%$ (App.~\ref{app:human}). 

\noindent\textbf{Stage 2. Trait Reasoning (Psychologist).} The \textit{Psychologist} receives the verified observations and produces, for each Big Five trait, a structured analysis containing a trait-level assessment (mapped from the GT scores in the First Impressions~\cite{escalante2020modeling} to five ordinal levels),
a reasoning chain citing cues as evidence, and a confidence-weighted rationale. 

\noindent\textbf{Stage 3. MCQ Generation (Examiner).} \label{sec:mcq} The \textit{Examiner} consumes the verified observations and Psychologist analyses and generates seven cue-grounding MCQs spanning a cognitive taxonomy (Table~\ref{tab:mcq_taxonomy}, Figure~\ref{fig:overview}) organized from \emph{reasoning} to \emph{visual grounding}. The \emph{reasoning cluster} probes higher-order social-cognitive abilities established in psychology and video QA: \emph{Personality Attribution}~\cite{funder1995accuracy} (behavior$\to$trait inference), \emph{Counterfactual} reasoning~\cite{roese1997counterfactual}, \emph{Temporal-Causal} chains~\cite{xiao2021nextqa}, and \emph{Mixed Emotion} discrimination~\cite{larsen2001happy}. The \emph{visual-grounding cluster} probes fine-grained perceptual localization: \emph{Micro-expression}~\cite{ekman1969nonverbal,yan2014casme} detection, \emph{Spatial Localization} of body regions~\cite{yu2016modeling,liu2024single}, and joint \emph{Temporal-Spatial} grounding~\cite{zhang2020vidstg,liu2025sfhand}. Each MCQ has six options: one correct answer and five distractors covering three failure modes (text-derivable, plausible-but-wrong-segment, near-miss). 

\noindent\textbf{Stage 4. Quality Assurance (Aligner).} The \textit{Aligner} performs automated quality assurance on the MCQs through two layers: deterministic code checks (timestamp range, bounding-box validity) and LLM-level semantic review (consistency between MCQ correct answers and the personality analyses; factual alignment with the observations). 
Full Aligner protocol in Appendix~\ref{app:prompts}. 
Cross-judge robustness validation via Claude 4.5/Gemini 2.5 confirms stable T2 ranking ($\rho \geq 0.92$, App.~\ref{app:judge}).

\noindent\textbf{Stage 5. Filtering and Expert Review (Human + Text-only LLMs).} Each MCQ passes through a two-step quality gate. \emph{(a) Text-leakage filter.} Every MCQ is answered by two text-only LLMs (GPT-4o-mini and Gemini~Flash) using \emph{only} the question stem and options (no video, no observations); items that \emph{both} LLMs answer correctly are flagged as transcript-derivable and dropped, ensuring every retained question requires multimodal grounding. \emph{(b) Expert review.} 
Trained expert annotators review the surviving MCQs from the video, providing the
final human correction and quality control.

\begin{figure}[t]
\centering
\includegraphics[width=\figframeworkscale\linewidth]{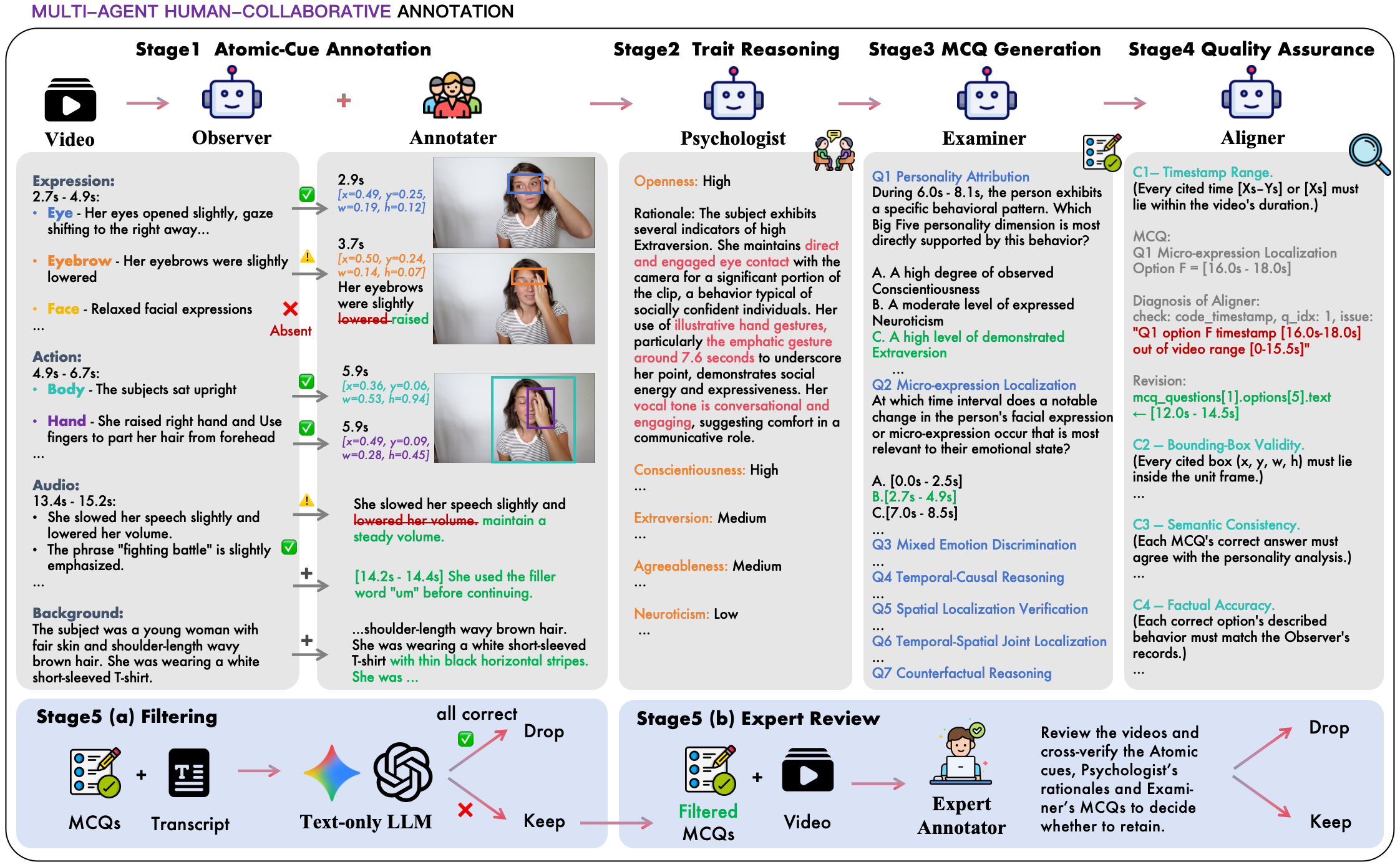}
\caption{\textbf{The five-stage multi-agent human-collaborative annotation pipeline.} Observer drafts atomic observations $\to$ Annotator verifies and localizes them (Stage 1) $\to$ Psychologist produces evidence-grounded Big Five analyses (Stage 2) $\to$ Examiner generates seven categories of cue-grounding MCQs (Stage 3) $\to$ Aligner enforces four consistency checks C1--C4 (Stage 4) $\to$ Stage 5 applies text-leakage filtering (a) and expert review (b).}
\label{fig:framework}
\vspace{-0.5cm}
\end{figure}

\vspace{-0.2cm}
\subsection{Dataset and Statistics}
\label{sec:data}
\vspace{-0.2cm}
\noindent\textbf{Source.}
\benchname{} draws its videos from the ChaLearn First Impressions V2 dataset~\cite{escalante2020modeling}, which contains $\sim$10K fifteen-second clips of single-person speech with crowd-sourced Big Five trait scores and ASR-extracted transcriptions. 

\noindent\textbf{Statistics.}
\label{sec:stats}
The released benchmark comprises 1{,}104 test videos accompanied by three layers of fine-grained annotations: $\sim$13.5K human-verified atomic behavioral observations 
across four perceptual channels (Expression, Action, Audio, Background); 5{,}520 trait-level personality analyses;
and 5{,}320 cue-grounding MCQs (averaging $4.8$ retained per video after filter). Continuous Big Five scores are discretized into the five ordinal levels of $\mathcal{L}$; the per-trait class distribution is reported in Appendix Table~\ref{tab:t1_dist}. Figure~\ref{fig:overview} jointly visualizes the resulting dataset structure.

\section{Evaluation Framework}
\label{sec:eval}

\benchname{} evaluates each model through three tasks of increasing cognitive depth (Figure~\ref{fig:framework}): \emph{ordinal personality rating} (T1), \emph{open-ended rating reasoning} (T2), and \emph{structured cue grounding} (T3); cross-task diagnostic rates (\S\ref{sec:cross_task}) then localize \emph{where} the personality-reasoning chain breaks.

\subsection{Task 1: Ordinal Personality Rating}
\label{sec:task1}
Given $V$, the model predicts $\hat{y}_i \in \mathcal{L}$ for each trait $i\!\in\!\mathcal{T}$ (Eq.~\ref{eq:t1}). Over a test set $\mathcal{D}_{\text{test}}$ of $N$ videos, we report exact-match accuracy and mean absolute error:
\begin{equation}
\operatorname{Acc}_{T1} = \frac{1}{5N}\sum_{n=1}^{N}\sum_{i\in\mathcal{T}} \mathbb{1}\!\left[\hat{y}_i^{(n)} = y_i^{(n)}\right],
\qquad
\operatorname{MAE}_{T1} = \frac{1}{5N}\sum_{n,\,i} \left|\hat{y}_i^{(n)} - y_i^{(n)}\right|,
\label{eq:t1_metrics}
\end{equation}
complemented by Spearman's $\rho$ in the appendix. Ordinal levels align with both human judgment and generative MLLM output formats better than continuous scores.

\subsection{Task 2: Open-Ended Rating Reasoning}
\label{sec:task2}

Given $V$, the model produces $(\hat{\mathcal{O}}, \hat{\mathcal{R}})$ (Eq.~\ref{eq:t2}): an open-ended explanation of \emph{why} the rating was given. An AI-as-Judge $J$ evaluates models output against GT along four dimensions: \emph{Evidence Coverage}, \emph{Logical Coherence}, \emph{Grounding Accuracy}, and \emph{Directional Accuracy}, collected in $\mathcal{D}_{\!J}$ with $|\mathcal{D}_{\!J}|=4$. Each dimension returns a score $s_d \in [1,10]$; we report the per-sample composite and its mean:
\vspace{-0.2cm}
\begin{equation}
S_{T2}(V, f_\theta) = \frac{1}{|\mathcal{D}_{\!J}|}\sum_{d \in \mathcal{D}_{\!J}} s_d\!\left(f_\theta(V);\, \text{GT}\right),
\qquad
\overline{S}_{T2} = \frac{1}{N}\sum_{n=1}^{N} S_{T2}(V_n, f_\theta).
\label{eq:t2_avg4}
\end{equation}
\vspace{-0.7cm}

\subsection{Task 3: Structured Cue Grounding}
\label{sec:task3}

Task~3 isolates the ability to \emph{ground personality judgments in specific observable cues} through structured multiple-choice probes. For each $q\!\in\!\mathcal{Q}_V$ in one of the seven cognitive categories $\mathcal{C}$ defined in Table~\ref{tab:mcq_taxonomy}, the model outputs $\hat{a}_q$ (Eq.~\ref{eq:t3}). We report overall and per-category accuracy:
\begin{equation}
\operatorname{Acc}_{T3} = \frac{1}{|\mathcal{Q}|}\sum_{q\in\mathcal{Q}} \mathbb{1}[\hat{a}_q = a_q^\star],
\qquad
\operatorname{Acc}_{T3}^{(c)} = \frac{1}{|\mathcal{Q}_c|}\sum_{q\in\mathcal{Q}_c} \mathbb{1}[\hat{a}_q = a_q^\star], \quad c\in\mathcal{C}.
\label{eq:t3_metrics}
\end{equation}
\vspace{-0.7cm}

\subsection{Cross-Task Diagnosis: Gaps and Failure Modes}
\label{sec:cross_task}\label{sec:failure_modes}

Beyond per-task accuracy, \benchname{}'s three tasks combine to reveal \emph{where} a model's personality-reasoning chain breaks. We define five quantities that jointly localize the failure: two population-level signals and four sample-level rates (three failures + one success).

\noindent\textbf{Population-level signals.} We rank all evaluated models on each task. The \gapname{} (RGM) of a model is its average T2/T3 rank minus its T1 rank; a large positive RGM flags a model that rates correctly without comparably grounded downstream support. To probe whether grounding has democratized at the same pace as overall capability, we also report the closed-vs-open frontier-mean (top-3 within each ecosystem) gap $\Delta_{Tk}\!=\!\overline{\operatorname{M}}_{Tk}^{\text{open}}\!-\!\overline{\operatorname{M}}_{Tk}^{\text{closed}}$ as a robust ecosystem-level snapshot. We refer to the field-wide phenomenon that \emph{most ``correct'' ratings come without grounded evidence} --- captured jointly by high $\overline{\operatorname{PR}}$, low $\overline{\operatorname{HR}}$, and within-model rating-vs-grounding rank disconnect $\operatorname{RGM}\!>\!0$ --- as the \emph{\gapinv{}} (\S\ref{sec:gap_inversion}).
\begin{align}
\operatorname{RGM}(m) &= \tfrac{1}{2}\!\left[\operatorname{rk}_{T2}(m)+\operatorname{rk}_{T3}(m)\right] - \operatorname{rk}_{T1}(m), \label{eq:rgm}\\[-1pt]
\Delta_{Tk} &= \overline{\operatorname{M}}_{Tk}^{\text{open}} - \overline{\operatorname{M}}_{Tk}^{\text{closed}}. \label{eq:delta}
\end{align}
\vspace{-0.4cm}

\vspace{-0.2cm}
\noindent\textbf{Sample-level failure modes.} Each prediction either succeeds or fails on three independent axes (rating, reasoning, cue retrieval), placing the outcome into one of $2^3\!=\!8$ cells. Four cells correspond to interpretable archetypes: \emph{Prejudice Rate} (PR; right rating, wrong cues), \emph{Confabulation Rate} (CR; right rating, incoherent reasoning), \emph{Integration-failure Rate} (IR; right cues, wrong rating), and \emph{Holistic-Grounding Rate} (HR; all three correct). Formally, we binarize each task outcome by a threshold $\theta_k$:
\vspace{-0.2cm}
\begin{equation}
r_k = \mathbb{1}[R_k \geq \theta_k],\;\;
R_1 = \tfrac{1}{|\mathcal{T}|}\sum_i \mathbb{1}[\hat{y}_i = y_i^\star],\;
R_2 = \tfrac{S_{T2}}{10},\;
R_3 = \tfrac{1}{|\mathcal{Q}_V|}\sum_q \mathbb{1}[\hat{a}_q = a_q^\star],
\label{eq:rk}
\end{equation}
with defaults $\theta_1{=}\theta_3{=}0.5$ (majority-correct) and $\theta_2{=}0.7$ (the $\geq\!7$ judge bucket; sensitivity in Appendix~\ref{app:threshold}); the four rates are then
\begin{align}
\text{PR}(m) &\!=\! \Pr[r_3\!=\!0\!\mid\! r_1\!=\!1], &
\text{CR}(m) &\!=\! \Pr[r_2\!=\!0\!\mid\! r_1\!=\!1], \label{eq:PR}\\[-1pt]
\text{IR}(m) &\!=\! \Pr[r_1\!=\!0\!\mid\! r_3\!=\!1], &
\text{HR}(m) &\!=\! \Pr[r_1\!=\!1 \!\wedge\! r_2\!=\!1 \!\wedge\! r_3\!=\!1]. \label{eq:HR}
\end{align}
PR/CR/IR are minimized; HR is capturing full three-tier success. 
A $3\!\times\!3\!\times\!3$ threshold sweep confirms that the HR ranking is stable ($\rho\!\geq\!0.92$ across all 27 combos; Appendix~\ref{app:threshold}).

\begin{table}[t]
\caption{\textbf{Main \benchname{} leaderboard (27 models, sorted by HR within each group).} T2: T2-Avg4 AI-as-Judge composite (1--10). HR/PR/CR/IR: holistic-grounding / prejudice / confabulation / integration-failure rates (\%) from Eqs.~(\ref{eq:PR}--\ref{eq:HR}) at default thresholds. RGM: rating--grounding misalignment, the rank-based per-model gap (Eq.~\ref{eq:rgm}). \best{Bold}: best per column; \second{underline}: second best.}
\vspace{0.1cm}
\label{tab:leaderboard}
\centering
\scriptsize
\setlength{\tabcolsep}{5pt}
\renewcommand{\arraystretch}{0.95}
\resizebox{\tableaderboardscale\linewidth}{!}{%
\begin{tabular}{l@{\hspace{0.4em}}l@{\hspace{1.0em}}cccc@{\hspace{1.0em}}cccc@{\hspace{1.0em}}c}
\toprule
& & \multicolumn{4}{c}{\textbf{Per-task accuracy}} & \multicolumn{4}{c}{\textbf{Failure-mode rates (\%)}} & \\
\cmidrule(lr){3-6}\cmidrule(lr){7-10}
\textbf{Model} & \textbf{Size} & \textbf{T1\,$\uparrow$} & \textbf{MAE\,$\downarrow$} & \textbf{T2\,$\uparrow$} & \textbf{T3\,$\uparrow$} & \textbf{HR\,$\uparrow$} & \textbf{PR\,$\downarrow$} & \textbf{CR\,$\downarrow$} & \textbf{IR\,$\downarrow$} & \textbf{RGM} \\
\midrule
\humanrow{11} \\
Random baseline    & -- & 20.0 & --   & --   & 16.7 & -- & -- & -- & -- & -- \\
\midrule
\propgroup{11} \\
Gemini 3 Flash~\cite{google2025gemini3flash}      & API & \best{64.1}   & \best{0.42} & \best{6.65}   & \second{66.5} & \best{33.5}   & 17.2          & 44.7          & \best{28.7}   & $+0.5$  \\
GPT-5.5~\cite{openai2025gpt55}               & API & 56.0          & 0.51        & \best{6.65}   & 66.4          & \second{28.0} & \second{15.5} & 46.4          & 36.5          & $-0.5$  \\
Gemini 3.1 Pro~\cite{google2025gemini31pro}  & API & \second{57.3} & \second{0.50} & \second{6.59} & \best{70.6} & 27.4          & \best{10.8}   & 53.2          & \second{33.4} & $+0.0$  \\
Gemini 2.5 Pro~\cite{comanici2025gemini}     & API & 50.1          & 0.61        & 6.40          & 65.2         & 20.0          & 16.9          & 52.3          & 48.6          & $-7.0$  \\
GPT-5.4~\cite{openai2025gpt54}               & API & 48.7          & 0.60        & 6.48          & 52.6         & 17.9          & 33.0          & 46.1          & 48.7          & $-9.5$  \\
Gemini 2.5 Flash~\cite{comanici2025gemini}   & API & 43.1          & 0.69        & 6.37          & 56.5         & 16.9          & 28.0          & \best{43.9}   & 59.0          & $-16.5$ \\
Claude Opus 4.6~\cite{anthropic2025claudeopus46}     & API & 50.2          & 0.59        & 6.50          & 49.7         & 16.8          & 40.7          & 48.4          & 46.1          & $-5.5$  \\
Claude Sonnet 4.6~\cite{anthropic2025claudesonnet46} & API & 46.6          & 0.63        & 6.37          & 45.6         & 12.5          & 46.1          & 47.4          & 54.3          & $-9.0$  \\
Claude Haiku 4.5~\cite{anthropic2025claudehaiku45}   & API & 50.6          & 0.57        & 6.51          & 41.0         & 12.4          & 55.9          & \second{44.6} & 46.4          & $-1.5$  \\
GPT-5.4-mini~\cite{openai2025gpt54mini}      & API & 49.6          & 0.59        & 6.38          & 40.5         & 11.3          & 55.4          & 49.4          & 48.2          & $-2.0$  \\
o4-mini~\cite{openai2025o4mini}              & API & 48.0          & 0.62        & 6.05          & 43.4         &  7.8          & 48.2          & 71.7          & 51.7          & $-5.0$  \\
GPT-4o~\cite{openai2024gpt4o}                & API & 53.3          & 0.55        & 6.03          & 31.9         &  4.5          & 69.7          & 75.7          & 38.3          & $+11.0$ \\
GPT-4o-mini~\cite{openai2024gpt4o}           & API & 47.6          & 0.62        & 5.44          & 17.5         &  0.3          & 87.9          & 95.2          & 45.5          & $+5.0$  \\
\midrule
\opengroup{11} \\
Qwen3.5-397B-A17B~\cite{qwen2025qwen3}      & 397B & 53.1 & 0.55 & 6.45 & 48.1 & 15.9 & 41.5          & 54.3          & 40.9        & $+0.0$  \\
Qwen3-VL-235B-A22B~\cite{bai2025qwen25vl}   & 235B & 51.5 & 0.58 & 6.39 & 44.2 & 12.8 & 47.0          & 56.4          & 43.8        & $+1.0$  \\
Qwen3-VL-30B-A3B~\cite{bai2025qwen25vl}     & 30B  & 55.8 & 0.52 & 6.34 & 43.0 & 12.4 & 52.4          & 58.8          & 38.8        & $+8.0$  \\
Gemma-4-31B-it~\cite{deepmind2025gemma4}       & 31B  & 55.7 & 0.52 & 6.02 & 57.0 & 11.3 & 29.8          & 75.4          & 39.2        & $+4.5$  \\
Llama-4-Maverick-FP8~\cite{meta2025llama4maverick} & 402B & 55.9 & 0.53 & 6.01 & 36.6 &  5.8 & 64.5          & 75.8          & 37.4        & $+14.0$ \\
GLM-4.6V~\cite{hong2025glm}                & 108B & 48.2 & 0.62 & 5.86 & 42.0 &  4.6 & 52.7          & 80.7          & 51.2        & $-1.0$  \\
MiMo-VL-7B-RL~\cite{li2025mimo}        & 7B   & 51.1 & 0.56 & 5.82 & 38.9 &  3.6 & 56.5          & 84.1          & 42.2        & $+8.0$  \\
Qwen3-VL-8B~\cite{bai2025qwen25vl}          & 8B   & 50.0 & 0.60 & 5.80 & 37.0 &  2.8 & 62.4          & 85.5          & 47.4        & $+5.0$  \\
Step3-VL-10B~\cite{huang2026step3vl}         & 10B  & 42.4 & 0.71 & 5.51 & 36.3 &  0.9 & 62.3          & 92.9          & 62.1        & $-5.5$  \\
MiniCPM-o 2.6~\cite{yao2024minicpmv}        & 8B   & 44.7 & 0.65 & 4.79 & 28.6 &  0.6 & 67.8          & 95.1          & 55.5        & $+1.5$  \\
Qwen2.5-Omni-7B~\cite{xu2025qwen3}          & 7B   & 43.8 & 0.66 & 5.10 & 27.8 &  0.5 & 79.9          & 95.8          & 55.6        & $+0.0$  \\
Qwen2.5-VL-7B~\cite{bai2025qwen25vl}        & 7B   & 45.1 & 0.65 & 4.67 & 23.9 &  0.1 & 86.3          & 98.9          & 58.9        & $+4.5$  \\
InternVL3-8B~\cite{chen2024internvl}         & 8B   & 43.8 & 0.65 & 4.84 & 26.4 &  0.0 & 75.6          & 99.8          & 54.9        & $+0.0$  \\
LLaVA-NeXT-Video-7B~\cite{liu2024llava15}  & 7B   & 36.0 & 0.87 & 1.94 & 16.7 &  0.0 & 82.3          & 100.0         & 62.9        & $+0.0$  \\
\bottomrule
\end{tabular}%
}
\vspace{-0.5cm}
\end{table}

\vspace{-0.2cm}
\section{Benchmarking Results}
\label{sec:exp}

\subsection{Models and Evaluation Protocol}
\label{sec:setup}

We evaluate 27 representative MLLMs spanning 12 families: GPT~\cite{openai2023gpt4,openai2024gpt4o,openai2025gpt5,openai2025gpt55,openai2025gpt54,openai2025gpt54mini,openai2025o4mini}, Gemini~\cite{anil2023gemini,reid2024gemini15,comanici2025gemini,google2025gemini3,google2025gemini3flash,google2025gemini31pro}, Claude~\cite{anthropic2024claude3,anthropic2025claudeopus46,anthropic2025claudesonnet46,anthropic2025claudehaiku45}, Qwen-VL~\cite{bai2023qwen,wang2024qwen2vl,bai2025qwen25vl,xu2025qwen3,qwen2025qwen3}, Gemma~\cite{deepmind2025gemma4}, Llama~\cite{meta2025llama4maverick}, GLM~\cite{glm2024chatglm,hong2025glm}, InternVL~\cite{chen2024internvl}, MiniCPM~\cite{yao2024minicpmv}, MiMo~\cite{li2025mimo}, Step~\cite{huang2026step3vl}, and LLaVA~\cite{liu2023llava,liu2024llava15}; 13 are proprietary and 14 are open-source, with the full list and parameter sizes in Table~\ref{tab:leaderboard}. We uniformly sample frames per video and use the same structured prompt per task for all models; open-source models are served via vLLM~\cite{kwon2023vllm}. For Task~2 we use GPT4o-mini as the AI-as-Judge, with a confidently-wrong consistency check in Appendix~\ref{app:judge}. A cross-judge robustness check with Claude Haiku 4.5 and Gemini 2.5 Flash-Lite confirms the T2 ranking is stable across judge families (Spearman $\rho \geq 0.92$, Appendix~\ref{app:judge}). Compute resources are detailed in Appendix~\ref{app:compute}.

\subsection{Leaderboard and the Prejudice Gap}
\label{sec:leaderboard}\label{sec:gap_inversion}

Table~\ref{tab:leaderboard} reports the full leaderboard, sorted by Holistic-Grounding Rate (HR). Our evaluation uncovers a pervasive \emph{\gapinv{}} across the 27 evaluated MLLMs, the mean Prejudice Rate is $\overline{\operatorname{PR}}\!=\!51.3\%$, where over half of correct ratings are ungrounded. Meanwhile, the mean Holistic-Grounding Rate is only $\overline{\operatorname{HR}}\!=\!10.4\%$, with the field's best model (Gemini~3~Flash) reaching just $33.5\%$. A traditional T1-only leaderboard would credit a model with $50$--$56\%$ rating accuracy as ``competent at personality assessment,'' yet on the same model, Prejudice Rate (PR) is typically $40$--$87\%$ (Table~\ref{tab:leaderboard}), most of those correct ratings rely on cues the model could not actually recover. Per-model PR-vs-T1 and PR/CR/IR/HR fingerprint visualizations are in Appendix~\ref{app:pr_t1} and \ref{app:fingerprint}.

The phenomenon is universal across the model landscape. Even at the proprietary frontier (Gemini~3~Flash, GPT-5.5, Gemini~3.1~Pro), Top-3 mean PR $\approx\!14.5\%$, leaving $1$ in $7$ correct ratings ungrounded; at the open-source frontier (Qwen3.5-397B, Qwen3-VL-235B, Qwen3-VL-30B), Top-3 mean PR $\approx\!47.0\%$. While the performance gap between open and closed frontiers remains narrow for rating ($\Delta_{T1}\!=\!-5.6\%$) and explanation ($\Delta_{T2}\!=\!-3.6\%$) , it widens for cue retrieval ($\Delta_{T3}\!=\!-26.6\%$; full table in Appendix~\ref{app:eco_gap}). Personality scoring and verbal reasoning have largely democratized; behavioral cue retrieval has not, and the open-source frontier is where prejudice is most prevalent. \S\ref{sec:diagnostics} drills into where this gap concentrates and how it interacts with per-sample failure modes.

\begin{figure}[t]
\centering
\begin{minipage}[c]{0.47\linewidth}
\centering
\includegraphics[width=\linewidth]{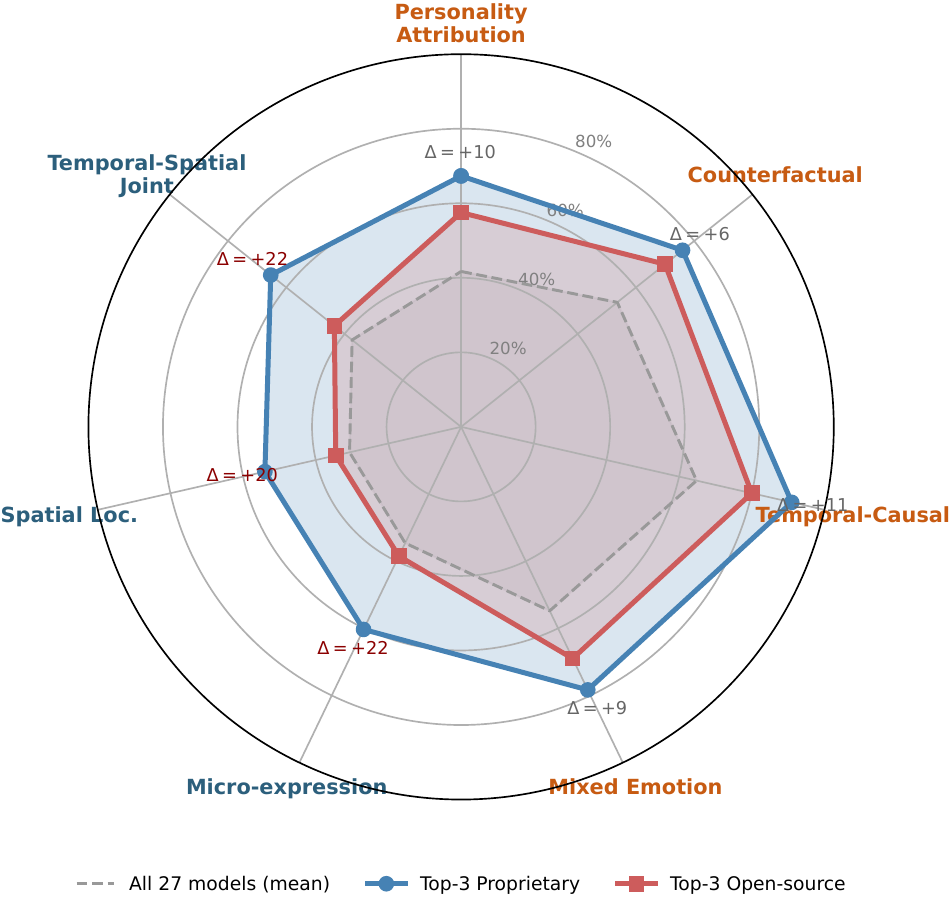}
\vspace{-0.6cm}
\captionof{figure}{\textbf{Per-category cognitive radar (T3).} Top-3 closed vs.\ Top-3 open accuracy across the seven cue-grounding MCQ categories. The closed-source advantage concentrates on the visual-grounding cluster (\emph{Spatial Localization}, \emph{Micro-expression}, \emph{Temporal-Spatial Joint}). 
}
\label{fig:radar}
\end{minipage}\hfill
\begin{minipage}[c]{0.50\linewidth}
\centering
\includegraphics[width=\linewidth]{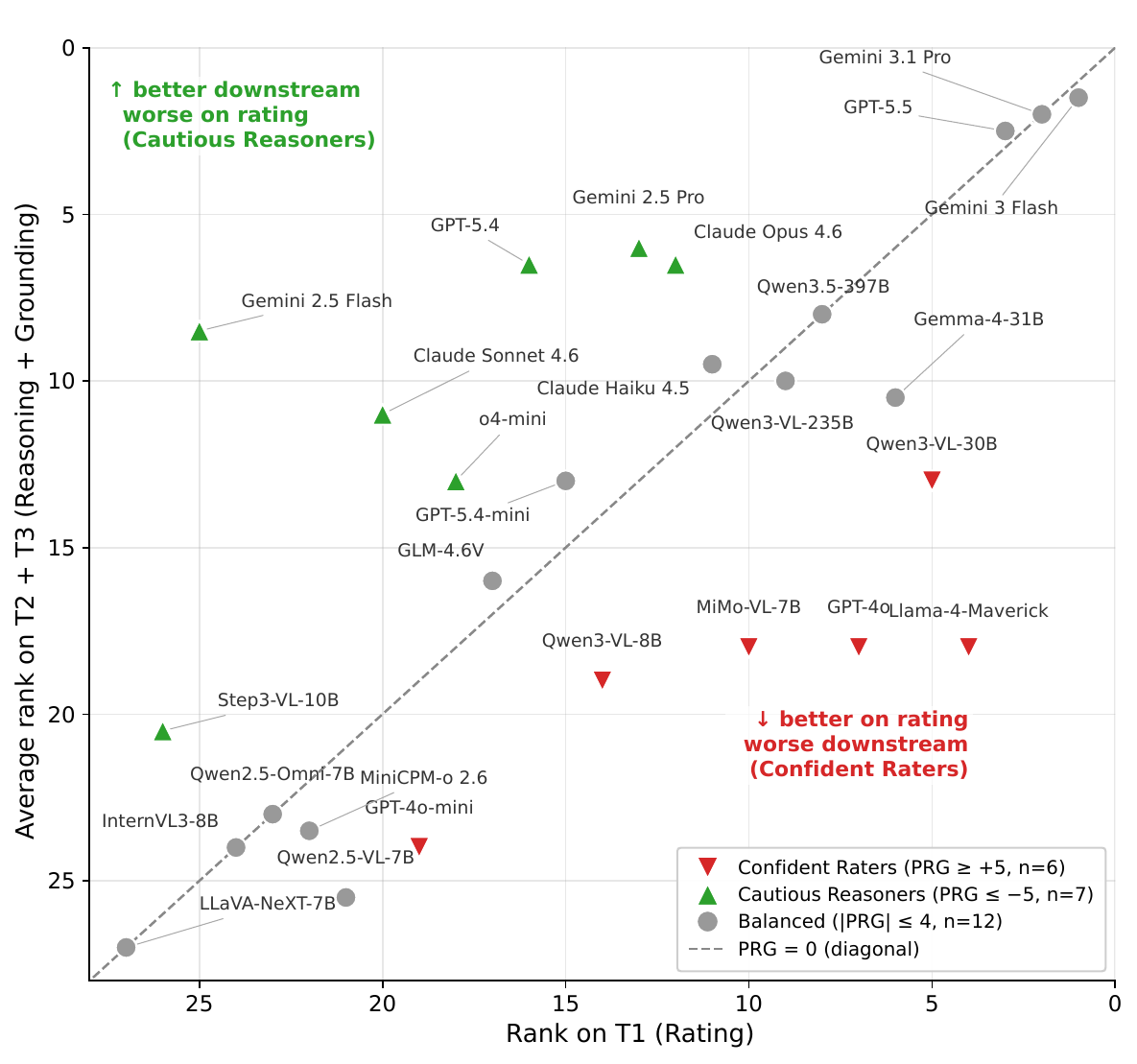}
\vspace{-0.6cm}
\captionof{figure}{\textbf{RGM archetypes scatter (T1 rank vs.\ avg.\ T2+T3 rank).} \emph{Confident Raters} (RGM\,$\geq\!+5$) lie above the diagonal: good on T1 but worse downstream. \emph{Cautious Reasoners} (RGM\,$\leq\!-5$) lie below: good downstream but rate poorly.
}
\label{fig:prg_archetypes}
\end{minipage}
\vspace{-0.5cm}
\end{figure}

\subsection{Where Prejudice Concentrates: Cognitive and Per-Sample Diagnostics}
\label{sec:diagnostics}\label{sec:category}\label{sec:failure_analysis}

We drill into the \gapinv{} along two complementary lenses: per-category cognitive sub-abilities (which T3 categories are systemic bottlenecks of cue retrieval) and per-sample diagnostic rates (which competence combinations break or succeed jointly).

\noindent\textbf{Per-category breakdown.} Mean accuracy across the 27 models reveals a stable difficulty hierarchy: \emph{Temporal-Causal Reasoning} is the easiest (64.8\%), while \emph{Spatial Localization} (30.7\%) and \emph{Micro-expression Localization} (34.6\%) are the hardest. The Top-3 closed advantage concentrates almost entirely on the visual-grounding cluster (Figure~\ref{fig:radar}), with $+19.5$\,pp on \emph{Spatial Localization} and $+21.8$\,pp on \emph{Temporal-Spatial Joint}, versus only $6$--$11$\,pp gaps on every reasoning-cluster category. Even the strongest closed model (Gemini~3.1~Pro) attains only 57\% on \emph{Spatial Localization} and 71\% on \emph{Temporal-Spatial Joint}, so fine-grained spatiotemporal grounding is a benchmark-wide bottleneck and the most actionable target for the next generation of open-source MLLMs. Full per-category accuracies are in Appendix~\ref{app:per_category} (Table~\ref{tab:per_category_mean}).

\noindent\textbf{HR as a highly discriminatory measure.} 
The PR/CR/IR/HR columns of Table~\ref{tab:leaderboard} (defined in \S\ref{sec:failure_modes}) decompose per-sample errors into interpretable archetypes. HR spans $0.0\%$ (LLaVA-NeXT, InternVL3) to $33.5\%$ (Gemini~3~Flash); its coefficient of variation $\text{CV}\!\approx\!0.93$ is far larger than any single-task metric (T1 $\approx\!0.13$, T2 $\approx\!0.16$, T3 $\approx\!0.36$), so conditioning on \emph{joint} rating--reasoning--grounding success amplifies the spread between models well beyond any individual accuracy. Across the 27 models, HR rankings remain strongly correlated with the equally-weighted task mean (Spearman $\rho\!\approx\!0.97$ vs.\ rank by $\bar{T}=(\text{T1}+\text{T2}/10+\text{T3})/3$). 
Informative exceptions exist, Gemma-4-31B-it ranks 5th by task mean but only $13.5$th by HR, indicating its T1 and T3 successes are distributed across \emph{different} videos rather than co-occurring per video. This is the very pattern that the HR conditional, and the broader PR/CR/IR cross-task taxonomy (\S\ref{sec:failure_modes}), is designed to expose.

\noindent\textbf{The failure profile separates two model archetypes.} Rating--Grounding Misalignment (RGM, Eq.~\ref{eq:rgm}) cleanly partitions models into two archetypes (Figure~\ref{fig:prg_archetypes}). \emph{Confident Raters} (RGM\,$\geq\!+5$, $n=5$) score well on T1 but fail downstream: Llama-4-Maverick-FP8 (RGM $+14$) ranks 4 on T1 but only 17/19 on T2/T3. \emph{Cautious Reasoners} (RGM\,$\leq\!-5$, $n=5$) exhibit the opposite pattern: Gemini~2.5~Flash (RGM $-16.5$) rates poorly (rank 25) but excels on T2 and T3. The remaining 17 models lie in the balanced middle band ($|\text{RGM}|\!\leq\!4$). The decomposition offers clear diagnostic utility, as confident raters need better grounding while cautious reasoners need better rating calibration.

\subsection{Additional Analyses}
\label{sec:additional}

Beyond the headline findings of \S\ref{sec:gap_inversion}--\ref{sec:diagnostics}, we run a set of auxiliary analyses (full results in the appendix) organized around three questions. To localize \emph{where the difficulty in \benchname{} lives}, we examine per-trait T1 accuracy across the five OCEAN dimensions (Appendix~\ref{app:per_trait}) and the per-dimension T2 score breakdown across the four AI-as-Judge axes (Appendix~\ref{app:t2_dim}); jointly these reveal which traits and which reasoning aspects are intrinsically hardest. To probe \emph{which model attributes correlate with strong \taskabbr{} performance}, we compare open-source models grouped by parameter scale (Appendix~\ref{app:scaling}), trace generation-over-time effects within each closed-source family (Appendix~\ref{app:gen_effects}), and report an observational comparison of reasoning-capable vs.\ non-reasoning subsets (Appendix~\ref{app:reasoning_effect}). To \emph{stress-test our methodology and benchmark integrity}, we measure T1 prediction-distribution calibration relative to ground truth (Appendix~\ref{app:t1_dist}), positional-bias $\sigma$ on the A--F option-letter distribution as a cheap cue-retrieval health signal (Appendix~\ref{app:posbias}), inter-model rank correlation on per-video task scores (Appendix~\ref{app:rank_corr}), and a confidently-wrong consistency check on the AI-as-Judge (Appendix~\ref{app:judge}).

\section{Discussion and Conclusion}
\label{sec:conclusion}

We introduce \taskname{} (\taskabbr{}) and \benchname{}, a multi-granularity benchmark requiring MLLMs to ground personality judgments in observable evidence. Evaluating 27 MLLMs reveals a pervasive \emph{\gapinv{}}: $51\%$ of correct ratings lack grounded evidence, and the mean Holistic-Grounding Rate (HR) is only $10.4\%$. These results show that traditional rating-only evaluations systematically overestimate competence by crediting ungrounded predictions. While proprietary and open-source models perform similarly on rating and explanation ($\Delta < 6\%$), a substantial $-26.6\%$ gap exists in cue retrieval. As a highly discriminative metric, HR reveals that reasoning-intensive models increasingly lead the field. Prioritizing fine-grained spatiotemporal grounding in post-training is therefore essential for developing the next generation of trustworthy, personality-aware MLLMs.

\noindent\textbf{Limitations and future work.}
\benchname{} focuses on apparent personality from short, single-speaker English video clips; throughout this work, this denotes the specific construct from First Impressions V2. We evaluate Task 2 reasoning quality via an AI-as-Judge protocol. 
Natural extensions include cross-cultural and multilingual videos, multi-judge ensembles for Task~2 reliability, and richer grounding operationalizations beyond MCQ-based cue retrieval. We hope \benchname{} catalyzes MLLMs that \emph{genuinely understand}, rather than merely judge, the people they observe; ethical considerations and responsible-use guidelines are discussed in Appendix~\ref{app:ethics}.

\bibliographystyle{plainnat}
\bibliography{references}

\newpage
\appendix
\setcounter{figure}{0}
\setcounter{table}{0}
\renewcommand{\thefigure}{A\arabic{figure}}
\renewcommand{\thetable}{A\arabic{table}}

\definecolor{tocdesc}{gray}{0.40}
\definecolor{tocgrouph}{RGB}{40,80,140}
\newcommand{\tocrow}[3]{\textbf{\S\ref{#1}~~#2} & \textcolor{tocdesc}{#3} \\}
\newcommand{\tocgroup}[1]{\addlinespace[6pt]\multicolumn{2}{@{}l}{\textcolor{tocgrouph}{\textsc{\textbf{#1}}}} \\[2pt]}

\section*{Appendix Contents}
\label{app:toc}

\begin{tcolorbox}[
  colback=gray!5, colframe=gray!50,
  arc=3pt, boxrule=0.5pt,
  left=6pt, right=6pt, top=6pt, bottom=6pt,
  title={\textbf{Appendix Overview}},
  fonttitle=\normalsize,
  coltitle=black,
  colbacktitle=gray!15,
]
\small
The appendix is grouped thematically. Cross-references in the main text use the labels below.

\vspace{4pt}
\renewcommand{\arraystretch}{1.22}
\begin{tabular}{@{}p{0.39\linewidth}p{0.55\linewidth}@{}}

\multicolumn{2}{@{}l}{\textcolor{tocgrouph}{\textsc{\textbf{Pipeline \& Dataset}}}} \\[2pt]
\tocrow{app:prompts}{Annotation Pipeline Prompts}{Observer / Psychologist / Examiner / Aligner contracts.}
\tocrow{app:human}{Human Annotation Protocol}{24-annotator team, web tool, quality metrics.}
\tocrow{app:mcq}{MCQ Examples per Category}{One worked example per of the seven categories.}
\tocrow{app:datasheet}{Dataset Documentation}{Datasheets-for-Datasets disclosure.}
\tocrow{app:ethics}{Ethics and Responsible Use}{Dataset bias and misuse caveats.}

\tocgroup{Task 1. Ordinal Personality Rating}
\tocrow{app:per_trait}{Per-Trait T1 Difficulty}{Per-Big-Five accuracy and MAE breakdown.}
\tocrow{app:t1_dist}{Full T1 Prediction Distribution}{Per-model calibration vs.\ ground truth.}
\tocrow{app:offn}{Off-by-$N$ Error Distribution}{Magnitude profile of T1 misranking.}

\tocgroup{Task 2. Open-Ended Rating Reasoning}
\tocrow{app:t2_dim}{T2 Per-Dimension Breakdown}{Per-model means on the four AI-as-Judge axes.}
\tocrow{app:t2_dist}{Logical-Coherence Distribution}{Per-model coherence-score histograms.}

\tocgroup{Task 3. Structured Cue Grounding}
\tocrow{app:per_category}{Per-Category Accuracy}{Full numerical breakdown over the seven categories.}
\tocrow{app:difficulty}{Question Difficulty Distribution}{Item-level difficulty histogram.}
\tocrow{app:specialization}{Additional Model-Level Analyses}{Expertise vectors, parameter efficiency, substitutability.}

\tocgroup{Failure Modes \& Diagnostics}
\tocrow{app:pr_t1}{PR vs.\ T1 Visualization}{Trustworthy-zone scatter ($5/27$ reach it).}
\tocrow{app:fingerprint}{Failure-Mode Fingerprint}{Per-model PR / CR / IR / HR heatmap.}
\tocrow{app:rank_slope}{T1 \(\to\) HR Rank Reordering}{Per-model rank-shift slope chart.}
\tocrow{app:guess_vs_grounded}{Right Rating With vs.\ Without Grounding}{Worked side-by-side example contrasting two models.}
\tocrow{app:eco_gap}{Closed-vs-Open Task Gap}{Frontier-mean task-level differences.}
\tocrow{app:threshold}{Threshold Sensitivity}{Robustness of PR/CR/IR/HR to threshold choice.}

\tocgroup{Ecosystem-Level Effects}
\tocrow{app:scaling}{Open-Source Size Scaling}{Task accuracy by parameter count.}
\tocrow{app:reasoning_effect}{Effect of Reasoning Capability}{Reasoning vs.\ non-reasoning subsets (observational).}
\tocrow{app:gen_effects}{Generation-over-Time Effects}{Per-family generation curves.}
\tocrow{app:rank_corr}{Inter-Model Rank Correlation}{Spearman $\rho$ across the Top-10 models.}

\tocgroup{Methodology}
\tocrow{app:posbias}{Positional Bias}{Option-letter skew $\sigma$ and its T3 correlation.}
\tocrow{app:judge}{AI-as-Judge Protocol}{Judge prompt, rubric, reliability, and cross-judge robustness.}
\tocrow{app:compute}{Compute Resources}{GPU hours, API costs, and overhead breakdown.}

\end{tabular}
\end{tcolorbox}

\newpage

\section{Annotation Pipeline Prompts}
\label{app:prompts}

This appendix documents the four LLM-agent prompts of the construction pipeline (\S\ref{sec:pipeline}). Full prompt text and JSON output schemas are released with the dataset; below we summarize each agent's contract.

\noindent\textbf{Stage 1 (a). Observer.}
\emph{Input}: a 15-second video and its transcription.
\emph{System role}: ``You are a non-interpretive behavior recorder. Record only what is observable; never explain why.''
\emph{Output schema}: a JSON list of atomic observations, one per indivisible behavioral event, each with the fields \texttt{obs\_id}, \texttt{dimension} $\in\!\{$Expression, Action, Audio, Background$\}$, \texttt{t\_start}, \texttt{t\_end}, \texttt{description} (factual, $\leq\!20$ words), and \texttt{body\_part} when applicable. The Observer is explicitly forbidden from making personality or affect claims; this is enforced by prompt instruction and validated via tag-vocabulary checking at parse time.

\noindent\textbf{Stage 2. Psychologist.}
\emph{Input}: a video, its transcription, and the human-verified atomic observation list from Stage~1.
\emph{System role}: ``You are an expert personality psychologist applying Funder's Realistic Accuracy Model.''
\emph{Output schema}: a JSON object with five \texttt{trait\_analysis} entries (one per OCEAN trait), each containing a \texttt{level} $\in\!\mathcal{L}$ mapped from the GT continuous score, an \texttt{evidence} list of cited \texttt{obs\_id}s ($\geq\!1$ required), and a \texttt{rationale} ($\leq\!100$ words) that links the cited cues to the trait. The grounding constraint is enforced at parse time: any analysis that fails to cite at least one valid \texttt{obs\_id} is rejected and re-queried.

\noindent\textbf{Stage 3. Examiner.}
\emph{Input}: the verified observations and Psychologist analyses.
\emph{System role}: ``You are an exam writer probing fine-grained social cognition.''
\emph{Output schema}: seven MCQs covering the seven categories of Table~\ref{tab:mcq_taxonomy}, each with a \texttt{question} stem, six labeled \texttt{options} (A--F), a \texttt{correct\_answer} letter, an \texttt{explanation} citing supporting \texttt{obs\_id}s, and a \texttt{distractor\_strategy} tag indicating which of three failure modes (text-derivable, plausible-but-wrong-segment, near-miss) each distractor exploits. The Examiner is required to use timestamp anchors (\texttt{t\_start}--\texttt{t\_end}) verbatim from the Observer pool when constructing temporal MCQs, and to use bbox coordinates verbatim from the Stage-1 verifier's bbox refinements when constructing spatial MCQs.

\noindent\textbf{Stage 4. Aligner.}
\emph{Input}: the seven Examiner-generated MCQs plus the upstream observations and analyses.
\emph{Operation}: a two-layer quality-assurance pipeline. \emph{Layer 1 (deterministic code checks)}: timestamp ranges within the video duration, bbox coordinates in $[0, 1]^4$, no out-of-vocabulary categories. \emph{Layer 2 (LLM semantic review)}: consistency between MCQ correct answer and the Stage-2 trait conclusion (e.g., a question about Extraversion High should not have a correct answer about Extraversion Low), and factual alignment between MCQ option text and the cited observation. The Aligner can only modify MCQ fields; it cannot edit observations or analyses, which are treated as immutable upstream artifacts. We log every Aligner correction together with the failed-check identifier; aggregate correction rates are reported in Appendix~\ref{app:human}.

\section{Human Annotation Protocol}
\label{app:human}

\noindent\textbf{Annotator team and training.} Stage~1 verification was performed by 24 trained annotators. Each annotator completed a training session covering Big Five trait definitions, the four Observer dimensions, common mistakes (e.g., conflating ``smiling'' with ``Agreeableness'' before recording the cue), and the bounding-box drawing tool. 1,633 unique videos were submitted, of which 199 were assigned to two independent annotators for quality monitoring. Stage~5 expert review was performed by a smaller panel of psychology-trained reviewers who adjudicated edge cases.

\begin{figure}[t]
\centering
\includegraphics[width=0.99\linewidth]{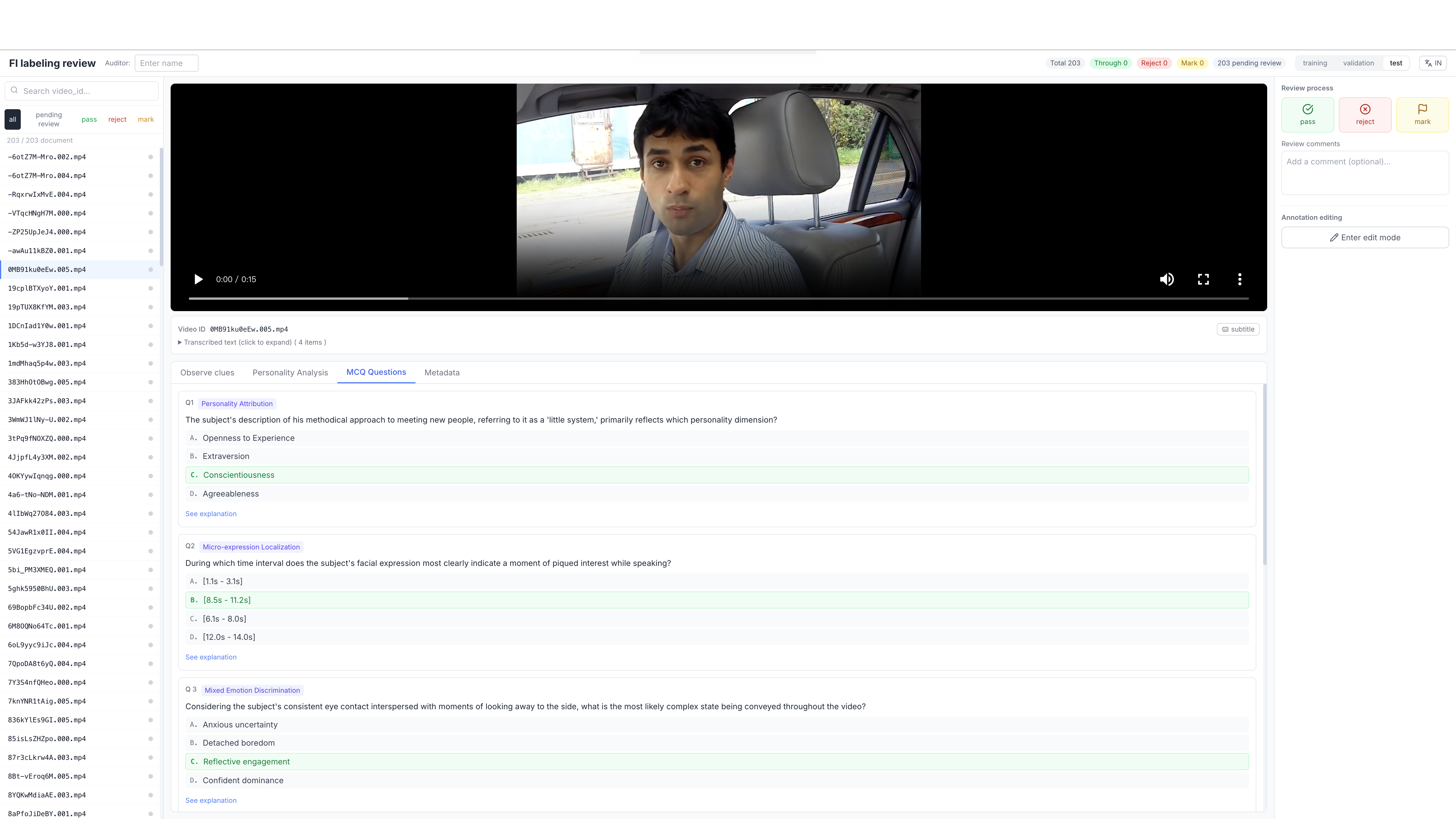}
\caption{\textbf{Annotation web tool.} Three-pane layout: frame-accurate video scrubber, atomic-cue list with edit controls, and bbox overlay on the current frame. Annotators verify each Observer-drafted cue, refine timestamps to frame precision, tighten bbox geometry, and label dimension / body-part. }
\label{fig:annotation_ui}
\end{figure}

\noindent\textbf{Web tool.} The annotation tool is a custom React application with three integrated views (Figure~\ref{fig:annotation_ui}): (i)~a frame-accurate video scrubber with $\leq\!1$-frame stepping, hotkeys for play/pause and $\pm\,1$-frame seek, and timestamp display in seconds and frames; (ii)~an OBS list view with editable per-cue timestamps, dimension dropdown, free-text description, body-part tag, and quality label (\emph{correct} / \emph{incorrect} / \emph{nonexistent}); (iii)~an overlay canvas for drawing tight bounding boxes on the current frame. Annotators draw timestamps and bounding boxes directly for every retained Expression or Action observation; the tool stores the original Observer draft alongside the annotator-corrected version for downstream auditing. JSON-schema validity is enforced on save.

\noindent\textbf{Per-stage decision rules.} For each Observer-drafted observation, the verifier selects one of three labels: \emph{correct} (kept verbatim), \emph{incorrect} (reworded), or \emph{nonexistent} (deleted). Incorrect / nonexistent rates by dimension are reported below. For every retained \emph{Expression} or \emph{Action} observation, the verifier additionally tightens the bbox and snaps the start/end timestamps to the closest perceptually-meaningful frame.

\noindent\textbf{Agreement and correction rates.}
Across the full annotation campaign, 24 annotators judged a total of $45{,}609$ Observer-drafted clues and drew $36{,}677$ bounding boxes. The aggregate quality breakdown is: $78.2\%$ of clues accepted as correct, $14.6\%$ corrected (wrong description, timestamp, or dimension), and $5.9\%$ deleted (nonexistent or hallucinated cue). Annotators also contributed 605 bonus clues (cues the Observer missed that annotators added).

\noindent\textbf{Annotator quality control.} Per-annotator acceptance rates varied from $66.9\%$ to $91.9\%$. Annotations with abnormally high acceptance rates were flagged during quality monitoring and filtered from the released dataset, removing approximately $8\%$ of submitted videos. The retained core annotators have a mean acceptance rate of $75.7\%$ and a combined correction-plus-deletion rate of $23\%$, demonstrating that human review adds substantial value beyond the LLM Observer.

\noindent\textbf{Inter-annotator agreement.} 199 videos were assigned to two independent annotators via a structured overlap pool. On the 147 pairs where both annotators submitted, the pairwise verdict agreement rate is $77.0\%$ (range $50$--$100\%$), confirming that the three-way verdict (correct / wrong / missing) is reproducible across annotators despite the inherent subjectivity of fine-grained behavioral cue assessment.
The Stage~5 panel reviews each MCQ that survives the text-leakage filter (Stage~5a). Across the released benchmark, $\sim\!7\%$ of post-filter MCQs receive an expert-corrected option, $\sim\!2\%$ receive a question-stem rewording, and $\sim\!1.5\%$ are dropped entirely.

\section{Full T1 Prediction Distribution}
\label{app:t1_dist}

Table~\ref{tab:t1_dist} reports each model's class distribution over the five ordinal levels and its total-variation distance to ground truth.

\begin{table}[t]
\caption{\textbf{Task~1 prediction distribution (\%) across the five ordinal levels (VL, L, M, H, VH), and total-variation distance (TVD) to ground truth (lower is better, 27 models).} Ground truth is listed first as a reference. Models follow the leaderboard order of Table~\ref{tab:leaderboard}. Open-source small models systematically collapse onto \emph{Medium}, while extreme classes (\emph{Very Low}/\emph{Very High}) are under-predicted by every model.}
\label{tab:t1_dist}
\centering
\small
\setlength{\tabcolsep}{7pt}
\renewcommand{\arraystretch}{1.12}
\scalebox{\tabtonescale}{%
\begin{tabular}{lrrrrrr}
\toprule
\textbf{Model} & \textbf{VL} & \textbf{L} & \textbf{M} & \textbf{H} & \textbf{VH} & \textbf{TVD} $\downarrow$ \\
\midrule
\humanrow{7} \\
\textbf{Ground truth} & \textbf{1.8} & \textbf{17.8} & \textbf{47.5} & \textbf{30.4} & \textbf{2.5} & -- \\
\midrule
\propgroup{7} \\
Gemini 3 Flash       &  0.4 & 18.1 & 43.3 & 37.5 &  0.7 &   7.4 \\
GPT-5.5              &  0.1 & 13.9 & 48.3 & 36.9 &  0.8 &   7.4 \\
Gemini 3.1 Pro       &  0.3 & 19.0 & 43.8 & 36.3 &  0.6 &   7.1 \\
Gemini 2.5 Pro       &  2.2 & 20.8 & 35.1 & 40.3 &  1.6 &  13.3 \\
GPT-5.4              &  0.0 & 15.6 & 47.0 & 36.5 &  0.8 &   6.1 \\
Gemini 2.5 Flash     &  1.2 & 19.9 & 28.5 & 48.9 &  1.5 &  20.6 \\
Claude Opus 4.6      &  0.2 & 20.2 & 44.9 & 34.1 &  0.5 &   6.1 \\
Claude Sonnet 4.6    &  0.1 & 18.7 & 46.1 & 34.2 &  0.9 &   4.7 \\
Claude Haiku 4.5     &  0.1 & 15.0 & 46.2 & 37.5 &  1.2 &   7.1 \\
GPT-5.4-mini         &  0.8 & 20.8 & 56.0 & 21.9 &  0.5 &  11.5 \\
o4-mini              &  0.3 & 21.3 & 46.7 & 31.2 &  0.5 &   4.3 \\
GPT-4o               &  0.1 & 16.9 & 47.5 & 33.6 &  1.9 & \best{3.3} \\
GPT-4o-mini          &  0.0 & 16.8 & 49.4 & 32.9 &  0.9 &   4.5 \\
\midrule
\opengroup{7} \\
Qwen3.5-397B-A17B    &  0.1 & 18.7 & 50.3 & 30.7 &  0.2 &   4.0 \\
Qwen3-VL-235B-A22B   &  0.1 & 17.5 & 42.2 & 39.4 &  0.7 &   9.1 \\
Qwen3-VL-30B-A3B     &  0.2 & 19.3 & 41.0 & 39.1 &  0.3 &  10.3 \\
Gemma-4-31B-it       &  0.6 & 16.8 & 48.2 & 33.0 &  1.4 & \second{3.4} \\
Llama-4-Maverick-FP8 &  0.2 & 23.9 & 38.5 & 37.1 &  0.4 &  12.7 \\
GLM-4.6V             &  0.2 & 22.3 & 43.1 & 34.1 &  0.3 &   8.3 \\
MiMo-VL-7B-RL        &  0.2 & 14.8 & 70.9 & 14.1 &  0.1 &  23.4 \\
Qwen3-VL-8B          &  0.3 & 24.4 & 40.9 & 33.0 &  1.4 &   9.2 \\
Step3-VL-10B         &  0.8 & 20.7 & 34.4 & 43.0 &  1.2 &  15.5 \\
MiniCPM-o 2.6        &  0.1 & 26.4 & 64.7 &  8.8 &  0.0 &  25.8 \\
Qwen2.5-Omni-7B      &  0.0 & 26.7 & 61.2 & 12.1 &  0.0 &  22.6 \\
Qwen2.5-VL-7B        &  0.1 & 22.6 & 60.3 & 17.0 &  0.0 &  17.6 \\
InternVL3-8B         &  0.2 & 20.1 & 72.6 &  7.0 &  0.1 &  27.4 \\
LLaVA-NeXT-Video-7B  &  0.0 & 63.4 & 19.5 & 17.0 &  0.0 &  45.6 \\
\bottomrule
\end{tabular}%
}
\end{table}

\section{Off-by-$N$ Error Distribution}
\label{app:offn}
Table~\ref{tab:offn} shows the distribution of T1 errors by absolute magnitude. The vast majority of errors ($\sim\!80\%$) are off-by-one across the 27 models, explaining why MAE is weakly discriminative on \benchname{} and why we primarily report accuracy.

\begin{table}[t]
\caption{\textbf{Task~1 error-magnitude distribution (\%) across 27 models.} \emph{Exact}: correct predictions; \emph{Off-$k$}: $|y-\hat{y}|=k$ on the five-level ordinal scale; \emph{Mean}: average absolute deviation. The vast majority ($\sim\!80\%$) of T1 errors are Off-1, so MAE is only weakly discriminative on \benchname{} and we emphasize exact-match accuracy in the main tables.}
\label{tab:offn}
\centering
\small
\setlength{\tabcolsep}{8pt}
\renewcommand{\arraystretch}{1.12}
\scalebox{\taboffnscale}{%
\begin{tabular}{lrrrrrr}
\toprule
\textbf{Model} & \textbf{Exact} $\uparrow$ & \textbf{Off-1} & \textbf{Off-2} & \textbf{Off-3} & \textbf{Off-4} & \textbf{Mean} $\downarrow$ \\
\midrule
\propgroup{7} \\
Gemini 3 Flash       & \best{63.4} & 30.6 &  5.6 &  0.4 &  0.0 & \best{0.43} \\
GPT-5.5              & 55.3 & 37.2 &  7.1 &  0.5 &  0.0 & 0.53 \\
Gemini 3.1 Pro       & 56.6 & 36.2 &  6.9 &  0.4 &  0.0 & 0.51 \\
Gemini 2.5 Pro       & 49.5 & 40.3 &  8.8 &  1.3 &  0.1 & 0.62 \\
GPT-5.4              & 48.0 & 42.8 &  8.7 &  0.5 &  0.0 & 0.62 \\
Gemini 2.5 Flash     & 42.7 & 45.3 & 10.8 &  1.1 &  0.0 & 0.70 \\
Claude Opus 4.6      & 49.6 & 41.2 &  8.5 &  0.7 &  0.0 & 0.60 \\
Claude Sonnet 4.6    & 45.9 & 44.2 &  9.2 &  0.8 &  0.0 & 0.65 \\
Claude Haiku 4.5     & 49.8 & 41.9 &  7.7 &  0.5 &  0.0 & 0.59 \\
GPT-5.4-mini         & 48.7 & 42.4 &  8.0 &  0.9 &  0.0 & 0.61 \\
o4-mini              & 47.2 & 42.9 &  9.1 &  0.8 &  0.0 & 0.64 \\
GPT-4o               & 52.6 & 39.3 &  7.5 &  0.7 &  0.0 & 0.56 \\
GPT-4o-mini          & 46.9 & 43.4 &  9.0 &  0.6 &  0.0 & 0.63 \\
\midrule
\opengroup{7} \\
Qwen3.5-397B-A17B    & 52.3 & 39.9 &  7.2 &  0.6 &  0.0 & 0.56 \\
Qwen3-VL-235B-A22B   & 50.8 & 39.9 &  8.5 &  0.8 &  0.0 & 0.59 \\
Qwen3-VL-30B-A3B     & 55.2 & 36.9 &  7.1 &  0.8 &  0.0 & \second{0.54} \\
Gemma-4-31B-it       & 54.8 & 37.2 &  7.3 &  0.7 &  0.0 & \second{0.54} \\
Llama-4-Maverick-FP8 & \second{55.3} & 35.6 &  8.3 &  0.8 &  0.0 & 0.55 \\
GLM-4.6V             & 47.6 & 42.2 &  9.3 &  0.9 &  0.0 & 0.63 \\
MiMo-VL-7B-RL        & 50.4 & 42.4 &  6.7 &  0.5 &  0.0 & 0.57 \\
Qwen3-VL-8B          & 49.6 & 41.0 &  8.7 &  0.8 &  0.0 & 0.61 \\
Step3-VL-10B         & 42.1 & 45.1 & 11.8 &  1.1 &  0.0 & 0.72 \\
MiniCPM-o 2.6        & 43.9 & 46.0 &  9.3 &  0.8 &  0.0 & 0.67 \\
Qwen2.5-Omni-7B      & 42.9 & 47.2 &  9.1 &  0.8 &  0.0 & 0.68 \\
Qwen2.5-VL-7B        & 44.3 & 46.0 &  8.9 &  0.8 &  0.0 & 0.66 \\
InternVL3-8B         & 43.0 & 48.0 &  8.3 &  0.7 &  0.0 & 0.67 \\
LLaVA-NeXT-Video-7B  & 35.8 & 42.8 & 19.3 &  2.1 &  0.0 & 0.88 \\
\bottomrule
\end{tabular}%
}
\end{table}

\section{Task~2 Logical-Coherence Distribution}
\label{app:t2_dist}

Table~\ref{tab:t2_dist} shows how each model's Task~2 \emph{Logical Coherence} scores distribute across five score buckets. Top models concentrate around the 8--9 bucket, while weaker models concentrate in 4--5. Virtually no model reaches the 10/10 bucket (only LLaVA-NeXT at $0.4\%$), confirming that our Judge does not inflate scores.

\begin{table}[t]
\caption{\textbf{Task~2 \emph{Logical Coherence} score distribution (\%) across five buckets, and per-model mean (27 models).} Top models concentrate $\sim\!50\%$ of samples in the 8--9 bucket, while weak models concentrate in 4--5. Virtually no model reaches the 10/10 bucket, confirming the Judge does not inflate scores. Models sorted within each group by mean.}
\label{tab:t2_dist}
\centering
\small
\setlength{\tabcolsep}{7pt}
\renewcommand{\arraystretch}{1.12}
\scalebox{\tabttwoscale}{%
\begin{tabular}{lrrrrrr}
\toprule
\textbf{Model} & \textbf{1--3} & \textbf{4--5} & \textbf{6--7} & \textbf{8--9} & \textbf{10} & \textbf{Mean} $\uparrow$ \\
\midrule
\propgroup{7} \\
Gemini 3 Flash       &  6.3 &  9.1 & 30.1 & \best{54.4} & 0.0 & \best{6.97} \\
GPT-5.5              &  5.2 & 10.6 & 32.2 & 51.9 & 0.0 & 6.94 \\
Gemini 3.1 Pro       &  6.5 & 11.2 & 29.4 & 53.0 & 0.0 & 6.91 \\
Gemini 2.5 Pro       & 10.3 & 11.5 & 29.2 & 49.1 & 0.0 & 6.70 \\
GPT-5.4              &  7.2 & 11.7 & 33.3 & 47.7 & 0.0 & 6.78 \\
Gemini 2.5 Flash     & 12.5 &  6.3 & 32.9 & 48.3 & 0.0 & 6.73 \\
Claude Opus 4.6      &  8.6 & 10.0 & 30.7 & 50.7 & 0.0 & 6.84 \\
Claude Sonnet 4.6    &  9.2 & 12.4 & 31.7 & 46.6 & 0.0 & 6.69 \\
Claude Haiku 4.5     &  7.3 & 11.1 & 30.8 & 50.9 & 0.0 & 6.89 \\
GPT-5.4-mini         &  8.7 & 10.7 & 34.7 & 45.9 & 0.0 & 6.65 \\
o4-mini              & 11.3 & 15.0 & 33.1 & 40.6 & 0.0 & 6.39 \\
GPT-4o               &  8.6 & 20.4 & 29.7 & 41.3 & 0.0 & 6.40 \\
GPT-4o-mini          & 11.4 & 26.8 & 41.5 & 20.3 & 0.0 & 5.86 \\
\midrule
\opengroup{7} \\
Qwen3.5-397B-A17B    &  7.6 & 10.6 & 32.8 & 48.9 & 0.0 & \second{6.81} \\
Qwen3-VL-235B-A22B   &  9.4 & 11.2 & 29.7 & 49.7 & 0.0 & 6.70 \\
Qwen3-VL-30B-A3B     & 10.4 & 12.1 & 29.4 & 48.1 & 0.0 & 6.62 \\
Gemma-4-31B-it       & 11.2 & 13.3 & 39.6 & 35.9 & 0.0 & 6.36 \\
Llama-4-Maverick-FP8 & 10.4 & 18.7 & 33.4 & 37.5 & 0.0 & 6.32 \\
GLM-4.6V             & 11.7 & 18.5 & 36.4 & 33.4 & 0.0 & 6.20 \\
MiMo-VL-7B-RL        & 10.3 & 21.4 & 40.4 & 28.0 & 0.0 & 6.14 \\
Qwen3-VL-8B          & 11.5 & 17.4 & 38.6 & 32.4 & 0.0 & 6.21 \\
Step3-VL-10B         & 14.1 & 19.8 & 39.5 & 26.6 & 0.0 & 5.96 \\
MiniCPM-o 2.6        & 25.2 & 27.5 & 32.6 & 14.6 & 0.0 & 5.17 \\
Qwen2.5-Omni-7B      & 19.4 & 27.3 & 37.7 & 15.6 & 0.0 & 5.43 \\
Qwen2.5-VL-7B        & 24.2 & 32.2 & 34.6 &  8.9 & 0.0 & 5.05 \\
InternVL3-8B         & 19.4 & 30.7 & 40.3 &  9.7 & 0.0 & 5.30 \\
LLaVA-NeXT-Video-7B  & 97.3 &  1.7 &  0.3 &  0.2 & 0.4 & 1.28 \\
\bottomrule
\end{tabular}%
}
\end{table}

\section{Question Difficulty Distribution}
\label{app:difficulty}
Across the 27 evaluated models, the number of models correctly answering a given question follows a bell-shaped distribution (Table~\ref{tab:difficulty}). The distribution's long left tail contains $153$ questions ($1.8\%$) that no model answered correctly, a ``human-only'' subset of \benchname{}.

\begin{table}[t]
\caption{\textbf{Number of models correctly answering each MCQ (out of 27) and the corresponding number of questions.} The distribution is roughly bell-shaped with a long left tail: 153 questions (1.8\%) are answered correctly by \emph{no} model.}
\label{tab:difficulty}
\centering
\small
\setlength{\tabcolsep}{14pt}
\renewcommand{\arraystretch}{1.15}
\scalebox{\tabdifficultyscale}{%
\begin{tabular}{cr@{\hskip 30pt}cr}
\toprule
\textbf{\# correct models} & \textbf{\# questions} & \textbf{\# correct models} & \textbf{\# questions} \\
\midrule
27 (all solved) & 134  & 13 & 285 \\
26              & 494  & 12 & 284 \\
25              & 456  & 11 & 322 \\
24              & 385  & 10 & 324 \\
23              & 373  &  9 & 284 \\
22              & 331  &  8 & 283 \\
21              & 380  &  7 & 287 \\
20              & 360  &  6 & 292 \\
19              & 334  &  5 & 293 \\
18              & 290  &  4 & 269 \\
17              & 304  &  3 & 263 \\
16              & 288  &  2 & 269 \\
15              & 275  &  1 & 178 \\
14              & 285  &  0 (human-only) & 153 \\
\bottomrule
\end{tabular}%
}
\end{table}

\section{Additional Model-Level Analyses}
\label{app:specialization}\label{app:efficiency}\label{app:substitutability}

\noindent\textbf{Relative expertise per model.}
Based on the per-category means in Table~\ref{tab:per_category_mean} and the radar in Figure~\ref{fig:radar}, for each model we compute its accuracy in each of the seven cue-grounding categories and subtract the model's own seven-category mean, yielding a centered \emph{expertise vector} that isolates relative strengths from absolute capability. Two robust patterns emerge across all 27 models. First, \emph{every} evaluated MLLM has positive deviation on the reasoning cluster and negative deviation on the visual-grounding cluster, confirming a field-wide preference for semantic reasoning over fine-grained perceptual localization. Second, individual signature strengths exist. Gemini~3.1~Pro is unusually strong on Temporal-Causal ($+27$\,pp above its own mean), Gemini~3~Flash on Personality Attribution ($+12$), and Gemma-4-31B-it leads the open-source field on Temporal-Causal ($+27$).

\begin{figure}[t]
\centering
\includegraphics[width=0.7\linewidth]{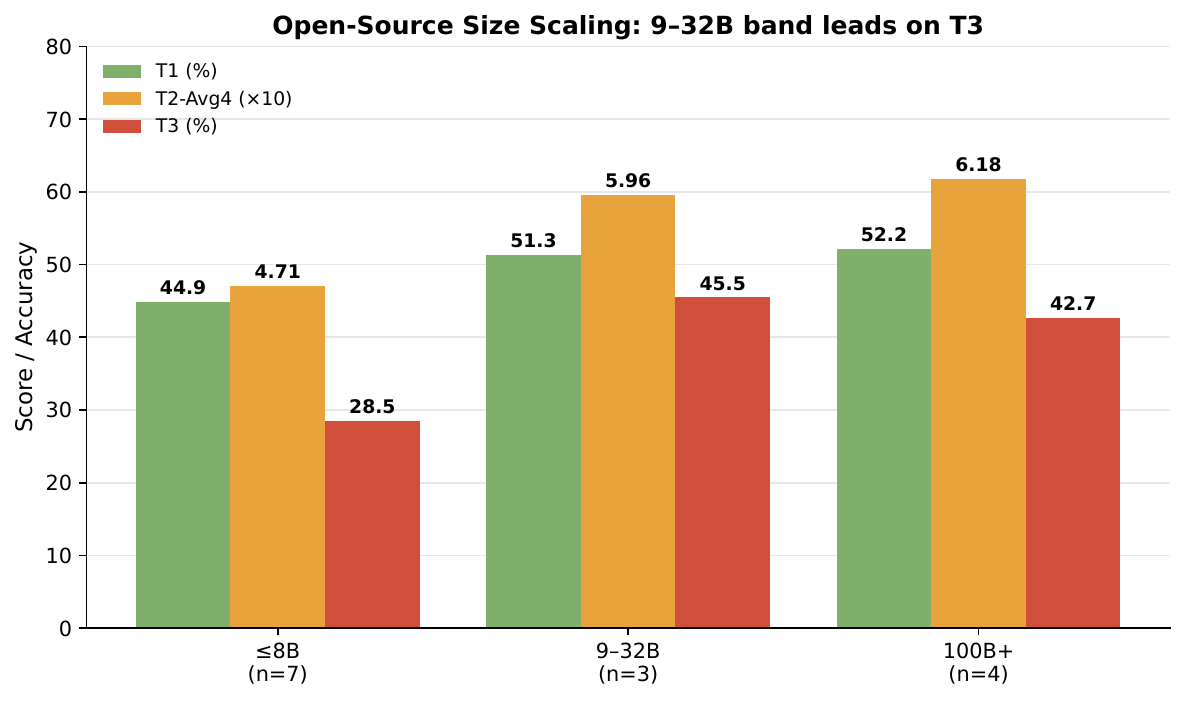}
\caption{\textbf{Open-source size scaling.} Each panel plots the per-band mean across $\leq\!8$\,B, $9$--$32$\,B, and $\sim\!100$\,B$+$. T3 plateaus past $\sim\!30$\,B while T1 and T2 keep rising slowly, suggesting cue-grounding requires post-training quality more than parameter count.}
\label{fig:scaling}
\end{figure}

\noindent\textbf{Parameter efficiency.}
Complementing Table~\ref{tab:scaling} and Figure~\ref{fig:scaling}, we measure parameter efficiency as billion-parameters per \%-T3-above-chance, where chance is $100\%/6\!\approx\!16.7\%$. The most efficient open-source model is MiMo-VL-7B-RL at $0.315$\,B per \%-T3-above-chance, followed by Gemma-4-31B-it ($0.77$) and Qwen3-VL-30B-A3B ($1.13$). At the heavy end, the 235--402\,B models cost $\sim\!8$--$10\!\times$ more parameters per percentage point. Data quality and post-training appear to matter more than parameter count.

\noindent\textbf{Substitutability matrix.}
Drawing on Table~\ref{tab:leaderboard}, we define open-source substitutability as the within-task gap between the strongest open-source model and the closed Top-3 median. \emph{T1 (rating)} is fully substitutable (gap $<\!2$\,pp). \emph{T2 (reasoning)} is partially substitutable (gap $\sim\!0.2$ on 10-pt scale). \emph{T3 (cue grounding)} is the bottleneck: only Gemma-4-31B-it ($57.0\%$) approaches the proprietary Flash class ($56.5\%$), and no open model reaches Flash-Pro tier ($65$+\%).

\section{MCQ Examples per Category}
\label{app:mcq}

We provide one worked example per of the seven cue-grounding categories, drawn from the released benchmark. All seven examples are from a single representative test video to allow cross-category comparison of the same behavioral context. Each item shows the question, all six options, the correct answer letter, and a one-line explanation. Full distractor strategies and OBS-ID evidence chains are in the released JSON files.

\begin{examplebox}[title={(1) Personality Attribution}]
\emph{Q:} During $11.6$\,s\,--\,$14.8$\,s, the person exhibits a specific behavioral pattern. Which Big Five personality dimension is most directly supported by this behavior?
\begin{itemize}[leftmargin=1.4em, itemsep=0pt, topsep=1pt, parsep=0pt]
\item[A.] High Conscientiousness, as the gesture reflects a planned and goal-oriented sequence.
\item[B.] High Extraversion, because the individual is demonstrating significant social energy.
\item[C.] High Agreeableness, due to the clear non-verbal effort to build rapport with the audience.
\item[D.] Low Openness, given that the subject is sharing a very common experience.
\item[E.] Medium Neuroticism, because his expression shows some on-camera tension.
\item[F.] Low Agreeableness, because his direct presentation could be interpreted as forceful.
\end{itemize}
\textcolor{humanc}{\textbf{Correct: C.}} \textcolor{gray}{ The leaning-in + widened eyes + raised brows during $11.6$--$14.8$\,s is a textbook rapport-building gesture, the strongest single cue for High Agreeableness in the clip.}
\end{examplebox}

\begin{examplebox}[title={(2) Counterfactual Reasoning}]
\emph{Q:} If the person had \emph{not} exhibited the behavior at $11.6$--$14.8$\,s, which personality assessment would be most affected?
\begin{itemize}[leftmargin=1.4em, itemsep=0pt, topsep=1pt, parsep=0pt]
\item[A.] Conscientiousness would seem higher, because the casual chewing-while-speaking would no longer be visible.
\item[B.] Agreeableness might rise from Medium to High, because this potentially assertive gesture is removed.
\item[C.] Extraversion would seem much lower, as his main moment of social engagement would be discounted.
\item[D.] Agreeableness might drop from High to Medium, as this key rapport-building gesture would be absent.
\item[E.] Openness would be most affected, since this gesture is the central part of his creative expression.
\item[F.] Neuroticism would increase, as the confident gesture's absence would emphasize earlier self-consciousness.
\end{itemize}
\textcolor{humanc}{\textbf{Correct: D.}} \textcolor{gray}{ Removing the rapport-building moment leaves no other strong evidence for High Agreeableness; the rating would conservatively drop one level.}

\end{examplebox}
\begin{examplebox}[title={(3) Temporal-Causal Reasoning}]
\emph{Q:} Which of the following best describes the causal chain linking the person's actions at different moments across the video?
\begin{itemize}[leftmargin=1.4em, itemsep=0pt, topsep=1pt, parsep=0pt]
\item[A.] A focused moment of assertion ($\sim$7\,s), a subsequent pause for reflection ($\sim$9\,s), then a cognitive withdrawal ($\sim$11\,s).
\item[B.] A cognitive withdrawal for formulation ($\sim$2\,s), a preparatory pause for structure ($\sim$5\,s), then a focused re-engagement for assertion ($\sim$7\,s).
\item[C.] An expressive social gesture ($\sim$5\,s), a period of anxious self-monitoring ($\sim$9\,s), and a compensatory return to neutrality ($\sim$13\,s).
\item[D.] A neutral opening ($\sim$1\,s), an emotional rapport display ($\sim$4\,s), then a logical conclusion to reinforce the argument ($\sim$12\,s).
\item[E.] A cognitive withdrawal ($\sim$2\,s), a focused re-engagement ($\sim$7\,s), and a concluding pause for planning ($\sim$11\,s).
\item[F.] An attempt at direct engagement ($\sim$1\,s), a disengaging gaze due to distraction ($\sim$5\,s), then a forced re-focus ($\sim$12\,s).
\end{itemize}
\textcolor{humanc}{\textbf{Correct: B.}} \textcolor{gray}{ The video shows look-away $\to$ pause $\to$ assertive re-engagement; B is the only option that matches the actual ordering of all three observed behaviors.}

\end{examplebox}
\begin{examplebox}[title={(4) Mixed Emotion Discrimination}]
\emph{Q:} During $1.9$--$8.4$\,s, the person's emotional state is best characterized as which of the following?
\begin{itemize}[leftmargin=1.4em, itemsep=0pt, topsep=1pt, parsep=0pt]
\item[A.] Engaged interest mixed with a feeling of clear surprise.
\item[B.] Subtle amusement tempered by professional composure.
\item[C.] Joyful excitement mixed with playful hesitation.
\item[D.] Quiet confidence tempered by a hint of genuine curiosity.
\item[E.] Mild anticipation mixed with focused readiness to act.
\item[F.] Calm satisfaction tempered with thoughtful recall.
\end{itemize}
\textcolor{humanc}{\textbf{Correct: E.}} \textcolor{gray}{ Setting up the action verbally (anticipation) plus tightened lips and gaze fixation (readiness to act) match E; the other options either over-claim affect (excitement, surprise) or claim emotions for which no cue is observable.}

\end{examplebox}
\begin{examplebox}[title={(5) Micro-expression Localization}]
\emph{Q:} At which time interval does a notable change in the person's facial micro-expression occur, most relevant to their High Agreeableness rating?
\begin{itemize}[leftmargin=1.4em, itemsep=0pt, topsep=1pt, parsep=0pt]
\item[A.] $[8.4\,\text{s}\,$--$\,9.6\,\text{s}]$
\item[B.] $[2.5\,\text{s}\,$--$\,3.7\,\text{s}]$
\item[C.] $[11.8\,\text{s}\,$--$\,13.0\,\text{s}]$
\item[D.] $[0.1\,\text{s}\,$--$\,1.3\,\text{s}]$
\item[E.] $[4.5\,\text{s}\,$--$\,5.7\,\text{s}]$
\item[F.] $[6.2\,\text{s}\,$--$\,7.4\,\text{s}]$
\end{itemize}
\textcolor{humanc}{\textbf{Correct: C.}} \textcolor{gray}{ The interval $[11.8, 13.0]$\,s captures the eye-widening + brow-raise transition that anchors the High Agreeableness rating; the other intervals contain only neutral baseline expression.}

\end{examplebox}
\begin{examplebox}[title={(6) Spatial Localization Verification}]
\emph{Q:} At around $12.6$\,s, a region was detected at $(x{=}0.296, y{=}0.657, w{=}0.093, h{=}0.091)$ (normalized, top-left origin). What is the most prominent expression or action in this region?
\begin{itemize}[leftmargin=1.4em, itemsep=0pt, topsep=1pt, parsep=0pt]
\item[A.] Subtly raised eyebrows signaling engagement.
\item[B.] An averted gaze directed well above the camera.
\item[C.] Jaw movement associated with active chewing.
\item[D.] A neutral facial expression showing no tension.
\item[E.] Slightly upturned mouth corners forming a smile.
\item[F.] A slightly opened mouth used for verbal emphasis.
\end{itemize}
\textcolor{humanc}{\textbf{Correct: F.}} \textcolor{gray}{ The provided bbox tightly frames the subject's mouth at $12.6$\,s, where he is leaning forward and speaking emphatically with the mouth slightly open.}

\end{examplebox}
\begin{examplebox}[title={(7) Temporal-Spatial Joint Localization}]
\emph{Q:} At which moment and location does the subject make a strong non-verbal gesture to emphasize a point to the viewer, involving a coordinated change in head position and facial expression?
\begin{itemize}[leftmargin=1.4em, itemsep=0pt, topsep=1pt, parsep=0pt]
\item[A.] $[2.6\,\text{s}]$ at $(x{=}0.52, y{=}0.49, w{=}0.18, h{=}0.40)$
\item[B.] $[2.8\,\text{s}]$ at $(x{=}0.39, y{=}0.58, w{=}0.13, h{=}0.09)$
\item[C.] $[0.2\,\text{s}]$ at $(x{=}0.67, y{=}0.88, w{=}0.17, h{=}0.12)$
\item[D.] $[13.8\,\text{s}]$ at $(x{=}0.22, y{=}0.56, w{=}0.19, h{=}0.29)$
\item[E.] $[10.4\,\text{s}]$ at $(x{=}0.37, y{=}0.50, w{=}0.17, h{=}0.27)$
\item[F.] $[4.9\,\text{s}]$ at $(x{=}0.38, y{=}0.41, w{=}0.15, h{=}0.08)$
\end{itemize}
\textcolor{humanc}{\textbf{Correct: D.}} \textcolor{gray}{ At $13.8$\,s the subject has leaned forward (peak emphatic gesture), and the bbox covers his face and upper torso at the moment the gesture peaks; the other options either point to neutral baseline timestamps or to too-tight bboxes that miss the head-coordinated motion.}
\end{examplebox}

\section{Failure-Mode Taxonomy: Threshold Sensitivity}
\label{app:threshold}

The per-sample binary outcomes $r_k=\mathbb{1}[R_k\!\geq\!\theta_k]$ defined in Eq.~(\ref{eq:rk}) depend on three thresholds $(\theta_1, \theta_2, \theta_3)$. Our default setting is $\theta_1\!=\!\theta_3\!=\!0.5$ (majority-correct: 3 of 5 traits on T1, 4 of 7 MCQs on T3) and $\theta_2\!=\!0.7$ (``acceptable'' judge quality, the lower bound of the $\geq\!7$ score bucket in Appendix Table~\ref{tab:t2_dist}). We assess robustness by sweeping each threshold independently in the ranges $\theta_1 \in \{0.4, 0.5, 0.6\}$, $\theta_3 \in \{0.4, 0.5, 0.6\}$, and $\theta_2 \in \{0.6, 0.7, 0.8\}$, and recomputing PR, CR, IR, HR from Eqs.~(\ref{eq:PR}--\ref{eq:HR}) for every combination.

\noindent\textbf{What stability looks like.} The qualitative claims of \S\ref{sec:failure_analysis} are robust if (i)~the Spearman rank correlation of HR with the default-threshold HR remains $\rho\!\geq\!0.9$ across the swept grid, (ii)~the Top-3 closed and Top-3 open identities are invariant, and (iii)~the sign of every $\Delta_{Tk}$ (\S\ref{sec:gap_inversion}) is preserved.

\noindent\textbf{Sweep results.} We sweep all $3\!\times\!3\!\times\!3\!=\!27$ threshold combinations and recompute HR for each of the 27 models. Across the full grid, the Spearman rank correlation of HR with the default-threshold HR is $\rho\!\in\![0.925, 1.000]$, confirming that the leaderboard ordering is highly stable. The Top-3 closed and open identities are preserved for 21 of the 27 combinations. The only regime where Top-3 changes is $\theta_2\!=\!0.8$ (requiring a Judge score $\geq\!8/10$), which collapses the field-best HR to $\sim\!4.7\%$ and makes rankings noisy. At $\theta_2\!\in\!\{0.6, 0.7\}$, the Top-3 is invariant regardless of $\theta_1$ or $\theta_3$.

The $\Delta_{Tk}$ closed-vs-open task-level gaps in Table~\ref{tab:gap_inversion} are computed from raw accuracy means, not from binarized rates, so they are unaffected by any threshold choice.

\noindent\textbf{Practitioner default.} We recommend $(\theta_1, \theta_2, \theta_3)\!=\!(0.5, 0.7, 0.5)$. Per-sample $R_k$ values will be released alongside the dataset so that future work can recompute the rates under any preferred threshold.

\section{AI-as-Judge Protocol}
\label{app:judge}

\noindent\textbf{Judge model.} For Task~2 we use \texttt{GPT-4o-mini} (temperature $0$, single sample) as a single-judge AI evaluator across all 27 evaluated MLLMs. We deliberately use a model that is \emph{not} on the leaderboard's high end to avoid self-preference bias.

\noindent\textbf{Four-dimension rubric.} The judge scores each per-trait Task~2 reasoning output on four independent axes, each in $[1, 10]$:

\begin{itemize}[leftmargin=1.3em, itemsep=1pt, topsep=2pt]
\item \textbf{Evidence Coverage}: does the rationale cite multimodal cues spanning the $\sim$15\,s clip, or does it rely on a single static impression?
\item \textbf{Logical Coherence}: do the cited cues genuinely entail the trait level, or is the explanation a non-sequitur?
\item \textbf{Grounding Accuracy}: are the cited cues observable in the video, or fabricated / generic?
\item \textbf{Directional Accuracy}: is the directional claim (high vs.\ low) consistent with the cited evidence?
\end{itemize}

The composite per-sample score $S_{T2}$ (Eq.~\ref{eq:t2_avg4}) is the simple mean of the four. The judge sees the model's Task~2 output, the ground-truth Big Five trait level, and the human-verified atomic observations as reference; it does not see the video.

\begin{figure}[t]
\centering
\includegraphics[width=0.7\linewidth]{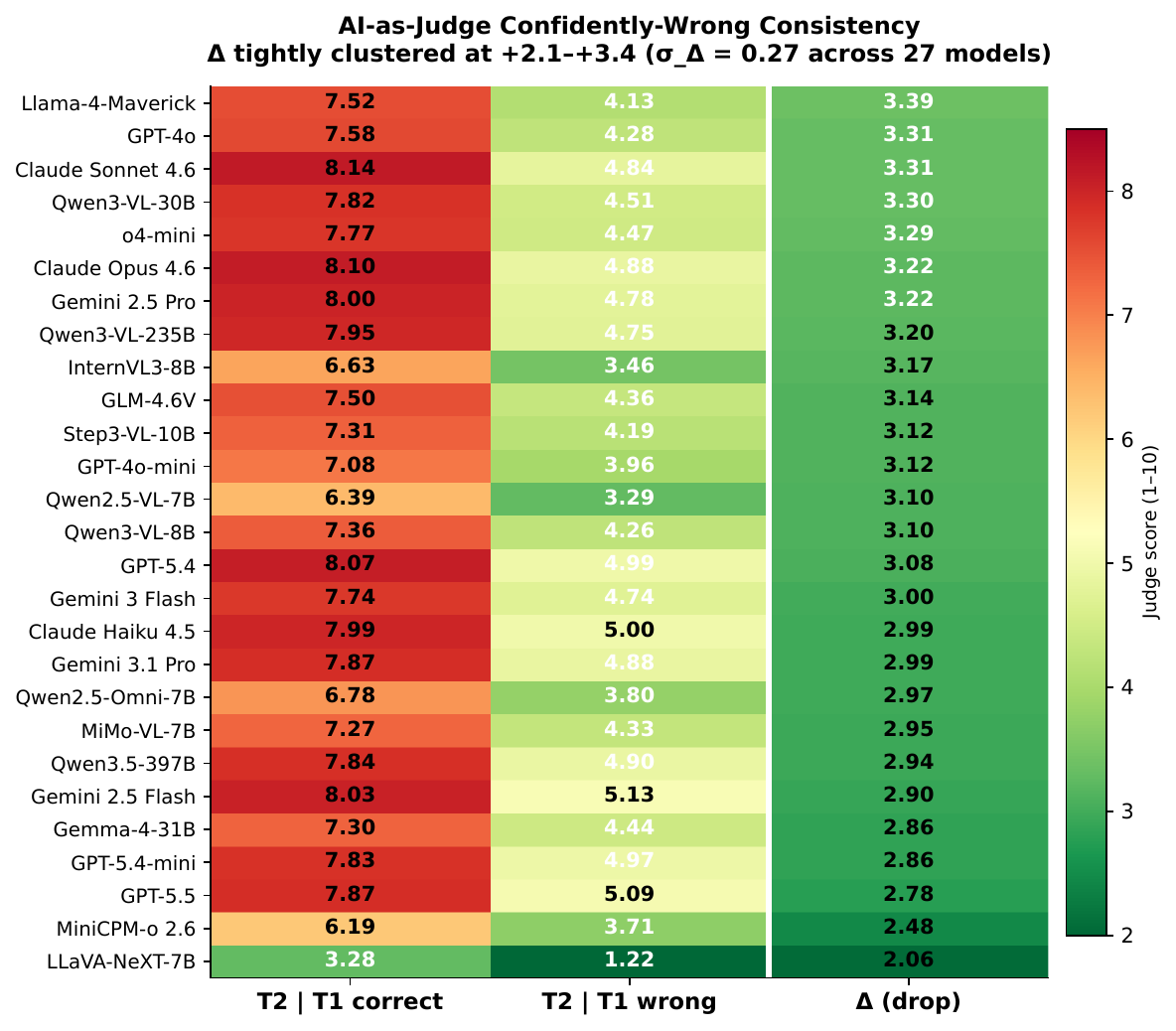}
\caption{\textbf{Confidently-wrong consistency of the AI-as-Judge.} Each row is a model: T2 mean given T1-correct (left), T2 mean given T1-wrong (middle), and the drop $\Delta$ (right). The $\Delta$ column is tightly clustered across all evaluated models, showing that the Judge tracks correctness rather than style.}
\label{fig:judge_reliability}
\end{figure}

\noindent\textbf{Confidently-wrong consistency check.} A standard robustness threat for AI-as-Judge is that it may simply mirror the surface ``style'' of the response rather than its correctness. We test this by partitioning each model's Task~2 outputs by whether Task~1 was \emph{correct} on that sample, and computing the conditional Judge mean for each partition (Figure~\ref{fig:judge_reliability}). Across all 27 evaluated models, the Judge gives systematically lower scores ($\Delta\!\approx\!2.1$--$3.4$ points) when T1 was wrong, even though Task~2 prose itself can look plausible to a casual reader. The cross-model standard deviation of this $\Delta$ is only $\sigma_\Delta\!=\!0.27$, indicating that the Judge applies the correctness-sensitive penalty uniformly rather than rewarding stylistic fluency.

\noindent\textbf{Robustness of AI-as-Judge evaluation.}
The tight across-model consistency of the confidently-wrong penalty (Figure~\ref{fig:judge_reliability}) indicates that Judge-based scoring in Task~2 is not simply rewarding style, but is tracking a latent correctness signal. Combined with the strong negative correlation between positional bias and MCQ accuracy (Figure~\ref{fig:posbias}), this suggests that \benchname{} is self-diagnostic: the same data that scores a model also reveals \emph{why} that model fails.

\noindent\textbf{Judge prompt skeleton.} The full judge prompt is in our supplementary material. The system message instructs the judge to be a strict psychology grader, the user message provides the model output, the GT trait level, and the human-verified observation list, and the response schema is a JSON object with the four dimension scores plus a one-sentence justification per dimension. We discard outputs that fail JSON-validation (rare; $<\!0.3\%$ of all samples) and re-query.

\noindent\textbf{Cross-judge robustness.} A reasonable concern with any single-judge protocol is that the leaderboard reflects judge-specific bias rather than genuine model differences, especially given that the GPT-4o-mini judge shares architectural lineage with two of the evaluated models (GPT-4o, GPT-4o-mini). To rule this out, we re-judge all 27 models' Task-2 outputs on a stable 200-video random subset (seed$\,=\,$42) with two alternative judges drawn from different families: \texttt{Claude Haiku 4.5} (Anthropic) and \texttt{Gemini 2.5 Flash-Lite} (Google), both comparable in capability tier to GPT-4o-mini. We use the identical prompt, rubric, and four-dimension scoring protocol.

Table~\ref{tab:cross_judge} summarizes the results. The Spearman rank correlations of $\overline{S}_{T2}$ between GPT-4o-mini and each alternative judge are $\rho\!=\!0.94$ ($p < 10^{-13}$) and $\rho\!=\!0.92$ ($p < 10^{-11}$) respectively, confirming that the T2 ranking is not an artifact of the judge's model family. The Top-3 identities by $\overline{S}_{T2}$ are preserved under both alternative judges.

A within-family check reveals that GPT-4o-mini scores its own family (GPT-4o, GPT-4o-mini) approximately $+1.0$ point higher on the 10-point scale than the cross-family judge average. This is a modest absolute inflation consistent with the known self-preference tendency of LLM judges, but it does not distort the relative ranking ($\rho \geq 0.91$). Haiku 4.5 scores $\sim\!0.7$ lower and Flash-Lite $\sim\!1.0$ lower than GPT-4o-mini on average across all models, a global calibration shift rather than a model-specific bias.

\begin{table}[t]
\caption{\textbf{Cross-judge robustness.} Left: Spearman $\rho$ between primary (GPT-4o-mini) and alternative judges. Right: within-family self-preference check for GPT-family models.}
\label{tab:cross_judge}
\centering
\small
\setlength{\tabcolsep}{6pt}
\renewcommand{\arraystretch}{1.15}
\begin{tabular}{lcc@{\hspace{2em}}lccc}
\toprule
\multicolumn{3}{c}{\textbf{Rank correlation}} & \multicolumn{4}{c}{\textbf{Within-family check}} \\
\cmidrule(lr){1-3}\cmidrule(lr){4-7}
\textbf{Judge pair} & \textbf{$\rho$} & \textbf{$p$-value} & \textbf{Model} & \textbf{GPT-4o-mini} & \textbf{Avg alt} & \textbf{$\Delta$} \\
\midrule
GPT-4o-mini vs Haiku 4.5 & 0.94 & $<10^{-13}$ & GPT-4o & 6.04 & 5.08 & $+0.97$ \\
GPT-4o-mini vs Flash-Lite & 0.92 & $<10^{-11}$ & GPT-4o-mini & 5.41 & 4.31 & $+1.10$ \\
\bottomrule
\end{tabular}
\end{table}

\section{Dataset Documentation}
\label{app:datasheet}

We document \benchname{} following the Datasheets for Datasets framework~\cite{gebru2021datasheets} (full Q\&A in the released artifact; key items below).

\noindent\textbf{Motivation.} The dataset was created to evaluate whether MLLMs can ground personality judgments in observable behavioral evidence (\S\ref{sec:intro}). It was created by the authors with no commercial sponsorship.

\noindent\textbf{Composition.} Each instance is a 15-second single-speaker English video clip drawn from ChaLearn First Impressions V2~\cite{escalante2020modeling}, paired with: (i) a transcription, (ii) human-verified atomic behavioral observations across four perceptual channels (Expression, Action, Audio, Background), (iii) five per-trait Big Five personality analyses with cited evidence, and (iv) seven cue-grounding multiple-choice questions covering the seven categories of Table~\ref{tab:mcq_taxonomy}. The released split contains $1{,}104$ videos, $\sim\!13.5$\,K verified observations, $5{,}520$ trait analyses, and $5{,}320$ MCQs ($4.8$ MCQs/video on average after the text-leakage filter). The Big Five labels are inherited from First Impressions V2 crowd-sourced annotations and discretized into five ordinal levels.

\noindent\textbf{Collection process.} Videos: drawn from First Impressions V2's existing test split. Annotations: produced by the multi-agent pipeline of \S\ref{sec:pipeline}. Stage~1 verification was performed by 24 trained annotators (1,633 unique videos submitted, 45,609 clues judged, 36,677 bboxes drawn) using the web tool described in Appendix~\ref{app:human}. Annotators were compensated at the local research-assistant hourly rate.

\noindent\textbf{Preprocessing / cleaning.} Beyond the multi-agent pipeline (Stages~1--5), we apply a text-leakage filter (Stage~5a) that drops any MCQ that two text-only LLMs (GPT-4o-mini and Gemini Flash) can both answer correctly from question + options alone. Videos retaining $<\!3$ MCQs after filtering are dropped from the released split.

\noindent\textbf{Uses.} Intended use is academic research on grounded personality reasoning and trustworthy multimodal evaluation. We discourage downstream deployment in personality screening or hiring without explicit consent and fairness audits (see Appendix~\ref{app:ethics}).

\noindent\textbf{Distribution.} Released under the same license as ChaLearn First Impressions V2 for the underlying videos plus a research-only license for our annotation layers, on \texttt{[release URL after acceptance]}.

\noindent\textbf{Maintenance.} The first author maintains the repository and a public issue tracker for label corrections. Errata releases are versioned and announced on the project page.

\section{Ethics and Responsible Use}
\label{app:ethics}

\noindent\textbf{Dataset bias.} \benchname{} inherits the cultural and linguistic biases of ChaLearn First Impressions V2~\cite{escalante2020modeling}, which is predominantly composed of Western-context English speakers. Personality perception is itself subjective and culturally situated; absolute trait scores should not be interpreted as objective ground truth across cultures.

\noindent\textbf{Misuse risks.} Automated personality assessment from video carries risks of misuse, including discriminatory screening (e.g., in hiring), surveillance, and over-claimed psychometric validity. \benchname{} is designed as a \emph{diagnostic research tool} for measuring grounded-reasoning capabilities of MLLMs, \emph{not} a deployment-ready personality-scoring system. We discourage downstream use without explicit consent, transparency about model limitations, and fairness audits of any system built on top of it.

\noindent\textbf{Operationalization caveats.} Task~3 operationalizes grounding as MCQ-based retrieval over a predefined cue set; a high Prejudice Rate may in part reflect MCQ-design choices (which cues are queried, which distractors are used) rather than a model's general inability to ground its judgment in observable behavior. The Task~2 AI-as-Judge, while scalable, may not capture every reasoning dimension; multi-judge or human-judge validation is recommended for high-stakes downstream use.

\section{T3 Per-Category Accuracy: Full Numerical Breakdown}
\label{app:per_category}

Table~\ref{tab:per_category_mean} reports the full per-category Task~3 accuracy referenced in \S\ref{sec:category} of the main paper. The left half lists the all-model mean and min--max range across the 27 evaluated MLLMs; the right half compares the Top-3 closed and Top-3 open averages and their absolute gap. The Mean column reproduces the all-model difficulty hierarchy referenced in the main text; the $\Delta$ column gives the precise per-category closed-vs-open gap visualized in Figure~\ref{fig:radar}.

\begin{table}[t]
\caption{\textbf{Task~3 per-category accuracy (\%).} \emph{Left}: mean across all 27 models with min--max range. \emph{Right}: Top-3 closed vs.\ Top-3 open averages and their absolute gap. The closed advantage concentrates on the visual-grounding cluster.}
\label{tab:per_category_mean}
\centering
\small
\setlength{\tabcolsep}{6pt}
\renewcommand{\arraystretch}{1.2}
\scalebox{\tabpercatscale}{%
\begin{tabular}{lcccccc}
\toprule
\textbf{Category} & \textbf{Mean (n=27)} & \textbf{Range} & & \textbf{Top-3 Closed} & \textbf{Top-3 Open} & \textbf{$\Delta$\,(pp)} \\
\midrule
\rowcolor{clusterbg}\multicolumn{7}{l}{\hspace{-0.3em}\textit{\textbf{Reasoning cluster}}}\\
Personality Attribution (Pers) & 41.7 & 15--70 & & 67.3 & 57.5 & $+9.8$ \\
Counterfactual (Counter)       & 53.6 & 15--80 & & 76.0 & 70.0 & $\mathbf{+6.0}$ \\
Temporal-Causal (TempC)        & \best{64.8} & 16--92 & & 91.0 & 80.0 & $+11.0$ \\
Mixed Emotion (Mixed)          & 54.8 & 15--80 & & 78.3 & 69.0 & $+9.3$ \\
\midrule
\rowcolor{clusterbg}\multicolumn{7}{l}{\hspace{-0.3em}\textit{\textbf{Visual Grounding cluster}}}\\
Micro-expression (Micro)       & 34.6 & 17--61 & & 60.3 & 38.5 & $+21.8$ \\
Spatial Loc.\ (Spat)           & 30.7 & 10--57 & & 54.0 & 34.5 & $+19.5$ \\
Temporal-Spatial Jnt.\ (TSJnt) & 37.4 & 13--71 & & 65.3 & 43.5 & $\mathbf{+21.8}$ \\
\bottomrule
\end{tabular}%
}
\end{table}

\section{Per-Trait T1 Difficulty}
\label{app:per_trait}

Table~\ref{tab:per_trait} reports mean T1 accuracy and MAE per Big Five trait, averaged across all 27 evaluated models. The trait-difficulty hierarchy is stable across the field: Extraversion / Agreeableness / Conscientiousness cluster at $53$--$55\%$; Openness sits slightly lower ($49\%$); Neuroticism is the universally hardest trait ($37.7\%$, MAE $0.87$). Even the strongest models (Gemini~3.1~Pro, Gemini~3~Flash, GPT-5.5) attain only $50$--$58\%$ on Neuroticism, suggesting that current MLLMs are bottlenecked by inferring internal emotional states from short ($\sim$15\,s) clips. Per-trait per-model details are in the trait-by-model heatmap in our supplementary material.

\begin{table}[t]
\caption{\textbf{Mean T1 accuracy and MAE per Big Five trait, averaged across all 27 models.}}
\label{tab:per_trait}
\centering
\small
\setlength{\tabcolsep}{14pt}
\renewcommand{\arraystretch}{1.15}
\scalebox{\tabpertraitscale}{%
\begin{tabular}{lcc}
\toprule
\textbf{Trait} & \textbf{Mean Acc (\%)} $\uparrow$ & \textbf{Mean MAE} $\downarrow$ \\
\midrule
Extraversion         & 54.5 & 0.51 \\
Agreeableness        & 53.8 & 0.52 \\
Conscientiousness    & 53.3 & 0.52 \\
Openness             & 49.2 & 0.58 \\
\midrule
Neuroticism & 37.7 & 0.87 \\
\bottomrule
\end{tabular}%
}
\end{table}

\begin{figure}[t]
\centering
\includegraphics[width=0.7\linewidth]{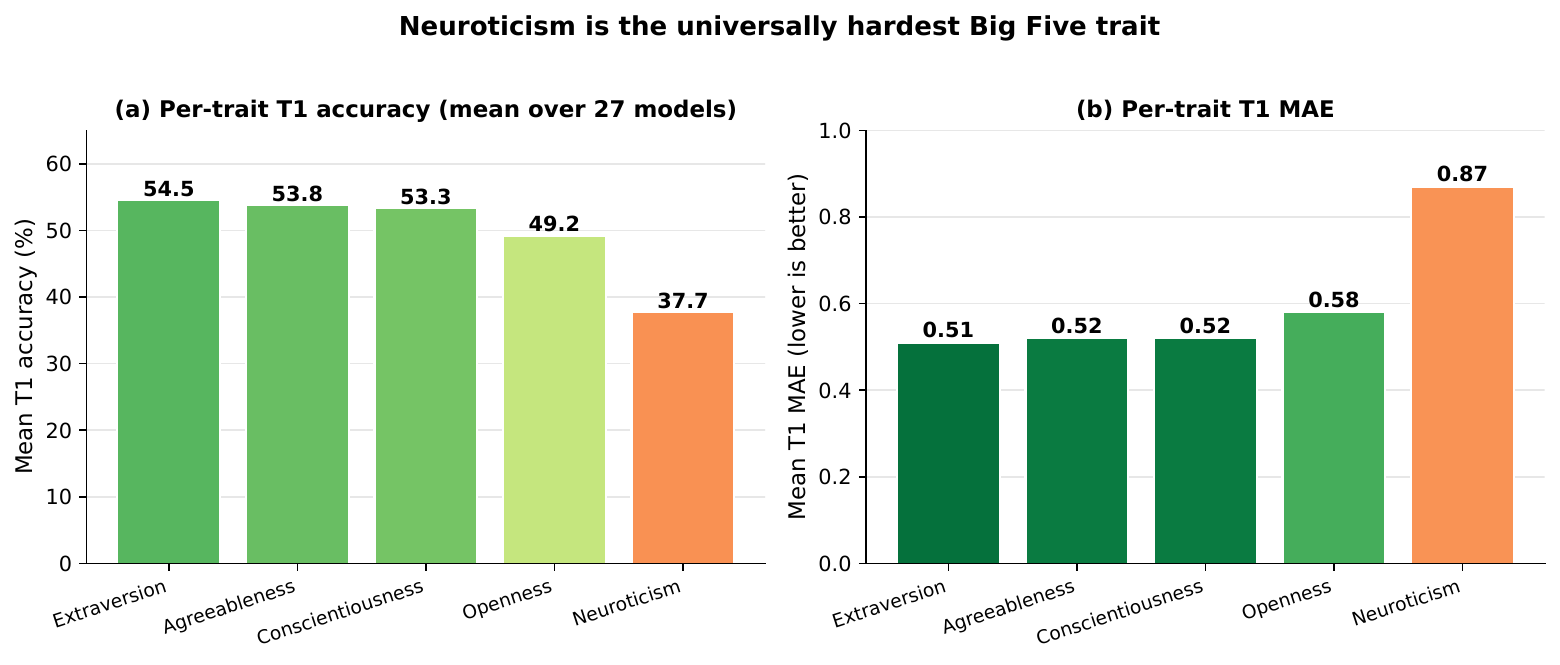}
\caption{\textbf{Per-trait T1 difficulty across the 27 evaluated MLLMs.} Neuroticism is the universally hardest trait; the gap from Openness ($49.2\%$) to Neuroticism ($37.7\%$) is larger than the gap from Extraversion to Openness, indicating that internal-state inference (Neuroticism) requires capabilities qualitatively beyond surface trait recognition.}
\label{fig:per_trait}
\end{figure}

\section{T2 Per-Dimension Breakdown}
\label{app:t2_dim}

The four AI-as-Judge dimensions exhibit different difficulty levels across the 27 models: Logical Coherence (mean $6.17$) and Grounding Accuracy (mean $6.43$) are easier; Directional Accuracy (mean $5.66$) sits in between; \emph{Evidence Coverage} (mean $5.14$) is the hardest. Models can write coherent on-topic explanations but tend to under-cite specific behavioral evidence, a finding consistent with the Confabulation Rate (CR) being the dominant failure pattern in mid-tier models (Table~\ref{tab:leaderboard}). A complete model-by-dimension heatmap is provided in the supplementary material.

\section{Open-Source Size Scaling}
\label{app:scaling}

Open-source MLLMs at three parameter scales (Table~\ref{tab:scaling}). Scaling from $\leq\!8$\,B to $9$--$32$\,B yields $+17$\,pp on T3, but scaling further to $100$\,B+ adds essentially nothing on T3 ($-2.8$\,pp due to Gemma-4-31B-it being the strongest open model on T3, beating the Qwen3-VL-235B-A22B and Llama-4-Maverick despite being $\sim$10$\times$ smaller). \emph{Sheer parameter count is not what limits open-source MCQ performance}; data quality and post-training appear to matter more.

\begin{table}[t]
\caption{\textbf{Open-source models grouped by parameter scale.}}
\label{tab:scaling}
\centering
\small
\setlength{\tabcolsep}{10pt}
\renewcommand{\arraystretch}{1.15}
\scalebox{\tabscalingscale}{%
\begin{tabular}{lcccc}
\toprule
\textbf{Scale} & \textbf{Count} & \textbf{T1 mean} & \textbf{T2-Avg4} & \textbf{T3 mean} \\
\midrule
$\leq 8$\,B    & 7 & 44.9 & 4.71 & 28.5 \\
$9$--$32$\,B   & 3 & 51.3 & 5.96 & \best{45.5} \\
$\sim 100$\,B$+$ & 4 & 52.2 & 6.18 & 42.7 \\
\bottomrule
\end{tabular}%
}
\end{table}

\section{Effect of Reasoning Capability (Observational)}
\label{app:reasoning_effect}

\noindent\textbf{Caveat.} The two subsets compared in this section differ on multiple confounding dimensions: parameter count, model family, generation, and training data. Reasoning-capable variants are typically newer, larger, and drawn from the strongest families. The aggregate gaps below should therefore be read as a \emph{descriptive field-level pattern} (``reasoning-capable variants tend to lead the leaderboard''), not as a controlled causal effect of explicit thinking. We report this analysis for completeness but do not headline it among our central findings.

\noindent\textbf{Setup.} We split the 27 evaluated MLLMs into a \emph{reasoning-capable} subset ($n\!=\!13$, models exposing an explicit thinking/reasoning mode) and a \emph{non-reasoning} subset ($n\!=\!14$), and compare the group means on each task and on HR.

\begin{table}[t]
\caption{\textbf{Reasoning-capable vs.\ non-reasoning models (observational).} Group means across the 27 evaluated MLLMs. The gap concentrates on T2/T3/HR, but the two subsets differ on size, family, and generation, so the gap is not a controlled effect of reasoning capability.}
\label{tab:reasoning}
\centering
\small
\setlength{\tabcolsep}{9pt}
\renewcommand{\arraystretch}{1.2}
\scalebox{\tabreasoningscale}{%
\begin{tabular}{lccccc}
\toprule
\textbf{Subset} & \textbf{T1\,(\%)} & \textbf{T2-Avg4} & \textbf{T3\,(\%)} & \textbf{HR\,(\%)} & \textbf{RGM\,(mean)} \\
\midrule
Reasoning ($n=13$)        & 51.0 & \best{6.35} & \best{52.2} & \best{16.4} & $-3.8$ \\
Non-reasoning ($n=14$)    & 48.5 & 5.38 & 33.9 & 4.9  & $+3.5$ \\
\midrule
$\Delta$ (Reasoning $-$ Non-reasoning) & $+2.6$ & $+0.97$ & $\mathbf{+18.3}$ & $\mathbf{+11.5}$ & $-7.3$ \\
\bottomrule
\end{tabular}%
}
\end{table}

\begin{figure}[t]
\centering
\includegraphics[width=0.85\linewidth]{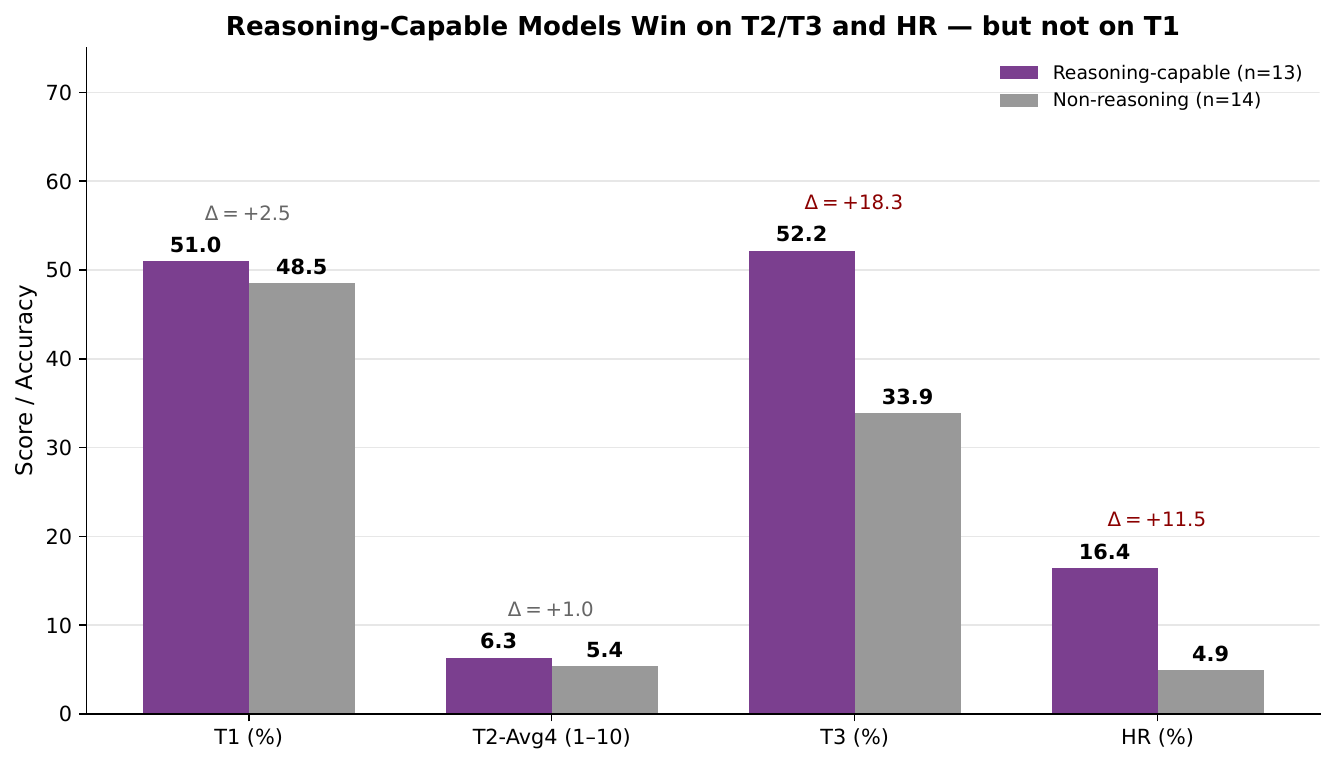}
\caption{\textbf{Reasoning-capable vs.\ non-reasoning subset means.} Differences are large on T3 / HR and small on T1, but the two subsets differ on size, family, and generation, so this gap is observational.}
\label{fig:reasoning_compare}
\end{figure}

\noindent\textbf{Observed pattern.} The reasoning-capable subset leads by $+18.3$\,pp on T3, $+11.5$\,pp on HR, and only $+2.6$\,pp on T1; mean RGM is also more negative ($-3.8$ vs.\ $+3.5$, Figure~\ref{fig:reasoning_compare}). Read together with Appendix~\ref{app:gen_effects}, the most parsimonious interpretation is that newer-generation, larger, top-family models (which happen to be reasoning-capable) dominate cue retrieval, while T1 saturates earlier and is therefore less sensitive to these confounded dimensions. We leave a controlled experiment (matched-size, matched-family, with vs.\ without thinking mode) to future work.

\section{Per-Model PR vs.\ T1 Visualization}
\label{app:pr_t1}

\begin{figure}[t]
\centering
\includegraphics[width=0.72\linewidth]{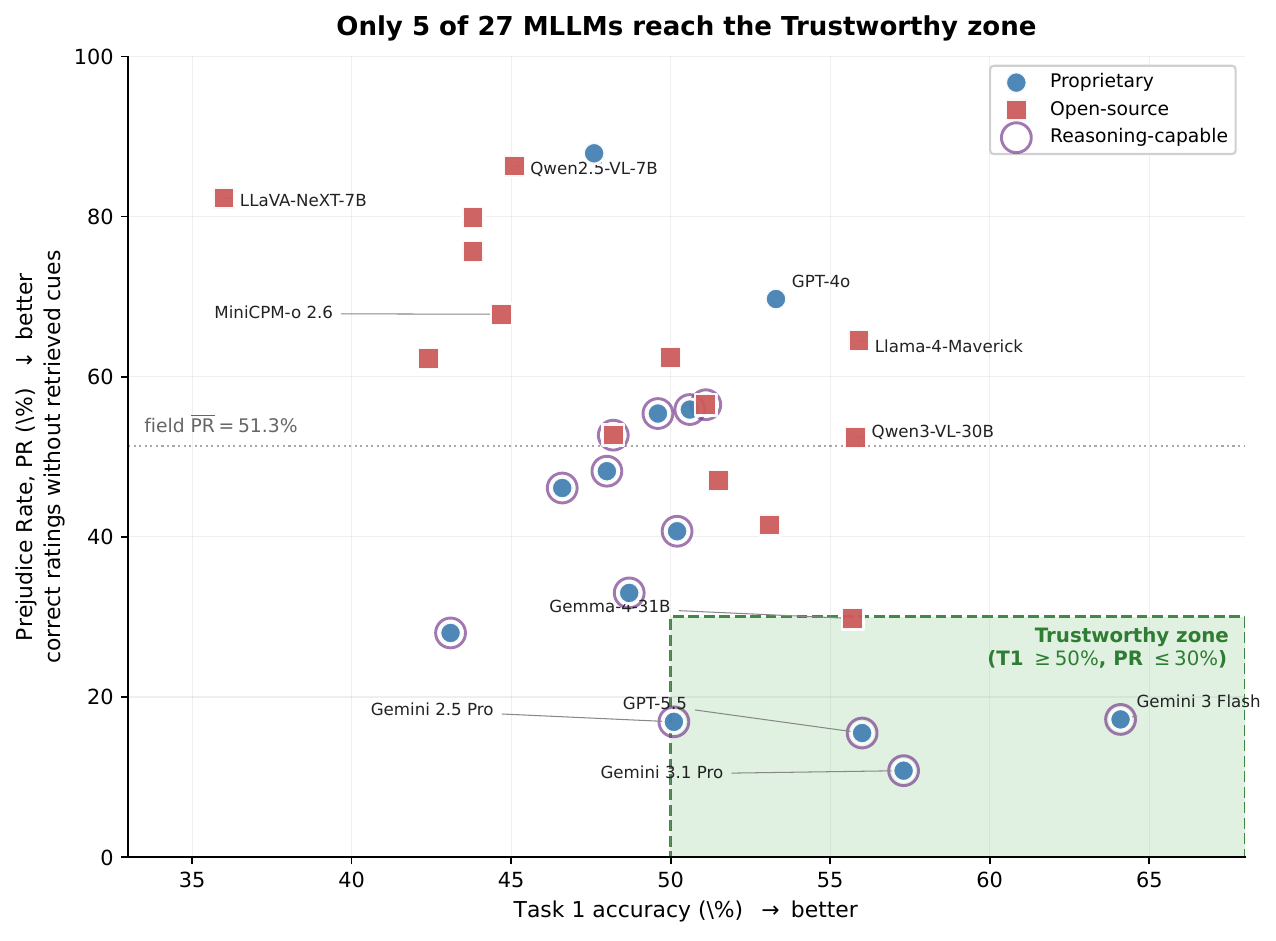}
\caption{\textbf{Right Rating, Wrong Cues.} T1 accuracy vs.\ Prejudice Rate across 27 MLLMs.}
\label{fig:pr_t1_scatter}
\end{figure}

Figure~\ref{fig:pr_t1_scatter} plots each evaluated MLLM as a point in (T1 accuracy, Prejudice Rate) space. The shaded \emph{Trustworthy zone} marks the desirable corner --- high T1 \emph{and} low PR. Only $5$ of $27$ models reach it (Gemini~3~Flash, GPT-5.5, Gemini~3.1~Pro, Gemini~2.5~Pro, and Gemma-4-31B-it, the only open-source model in the zone). The rest of the field clusters above the $\overline{\operatorname{PR}}\!=\!51.3\%$ reference line, confirming that the \gapinv{} is a field-wide rather than archetype-specific phenomenon.

\section{Per-Model Failure-Mode Fingerprint}
\label{app:fingerprint}

\begin{figure}[t]
\centering
\includegraphics[width=0.65\linewidth]{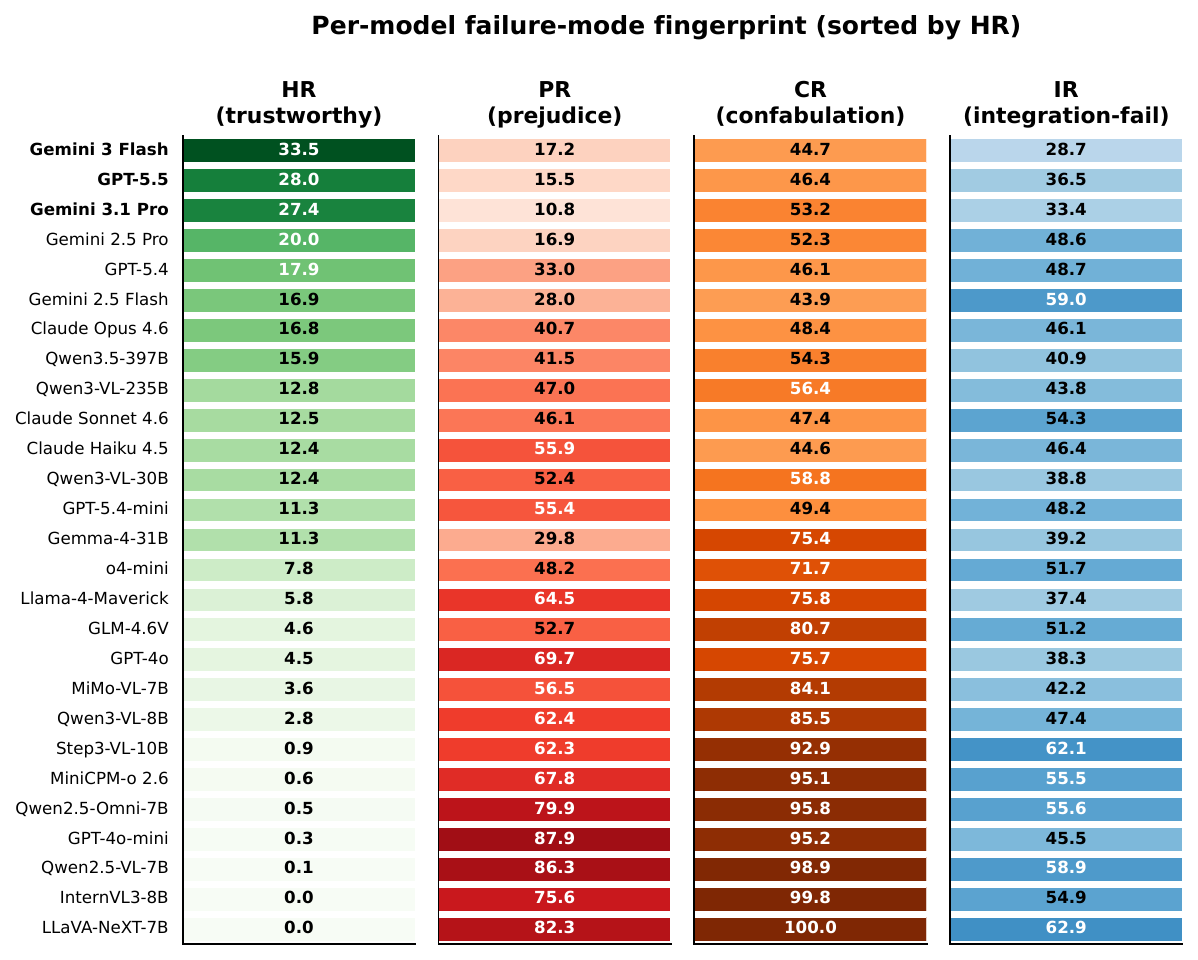}
\caption{\textbf{Per-model failure-mode fingerprint} (sorted by HR).}
\label{fig:failure_fingerprint}
\end{figure}

Figure~\ref{fig:failure_fingerprint} renders the four sample-level failure rates from Eqs.~(\ref{eq:PR}--\ref{eq:HR}) as a per-model heatmap, sorted by HR. Compared to scanning the same numbers in Table~\ref{tab:leaderboard}, the heatmap makes the field-wide pattern visually immediate: HR (green, ``trustworthy'') is concentrated almost entirely in the top three rows; PR (red, ``prejudice'') and CR (orange, ``confabulation'') saturate for the field's mid-to-bottom; IR (blue, ``integration-failure'') is broadly elevated across the table.

\section{T1 \(\to\) HR Rank Reordering}
\label{app:rank_slope}

\begin{figure}[t]
\centering
\includegraphics[width=0.55\linewidth]{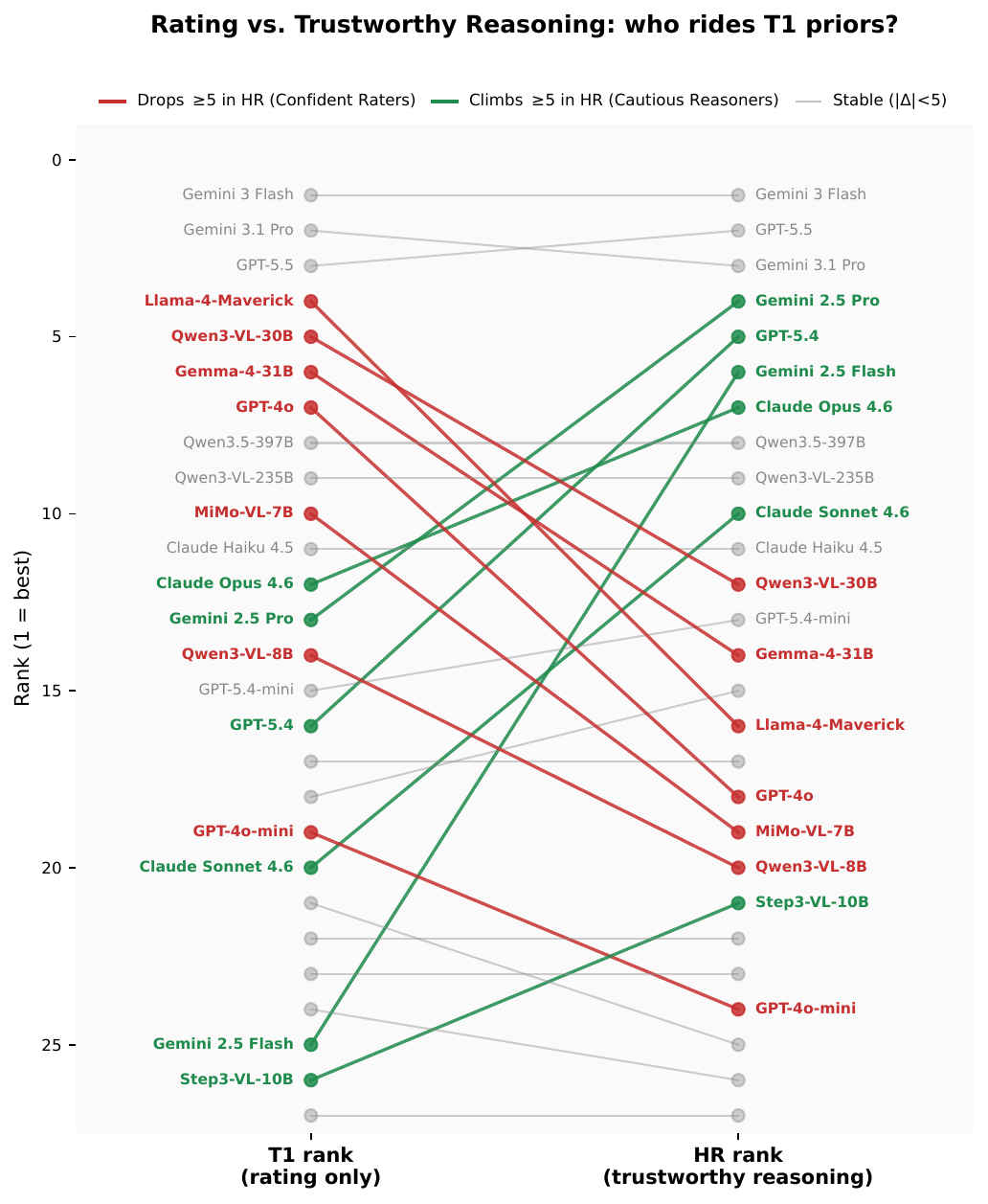}
\caption{\textbf{Rank reordering from T1 to HR.} Red = drops $\geq\!5$ in HR (\emph{Confident Raters}), green = climbs $\geq\!5$ in HR (\emph{Cautious Reasoners}), gray = stable.}
\label{fig:rank_slope}
\end{figure}

The Failure-Mode Fingerprint (Figure~\ref{fig:failure_fingerprint}) and the PR-vs-T1 scatter (Figure~\ref{fig:pr_t1_scatter}) both visualize \emph{rates}. Figure~\ref{fig:rank_slope} provides a complementary \emph{rank-based} view of the same phenomenon: for each model, we draw a line from its T1 rank (left, rating-only leaderboard) to its HR rank (right, trustworthy-reasoning leaderboard). Models that drop $\geq\!5$ ranks under HR are colored red (\emph{Confident Raters}); models that climb $\geq\!5$ ranks are colored green (\emph{Cautious Reasoners}); models with $|\Delta\text{rank}|\!<\!5$ are gray. The picture confirms what RGM (Eq.~\ref{eq:rgm}) measures: a small handful of models reorder substantially when HR replaces T1 as the ranking criterion. Llama-4-Maverick-FP8 (T1 rank 4 $\to$ HR rank 17) and GPT-4o (rank 5 $\to$ rank 18) drop the most, while Gemini~2.5~Flash (rank 24 $\to$ rank 5) and GPT-5.4 (rank 17 $\to$ rank 5) climb the most. Most models stay close to the diagonal, indicating that for the bulk of the field rating success and trustworthy reasoning are coupled.

\section{Worked Example: Right Rating With vs.\ Without Grounding}
\label{app:guess_vs_grounded}

To make the prejudice phenomenon concrete, we contrast two models on the same video, the same trait, and the same correct rating. The video is \texttt{W4tz3plvvKI.001} (a young man discussing his current and aspirational video-editing software); the trait is \emph{Extraversion} with ground-truth level \emph{Low}. Both GPT-4o and Gemini~3~Flash predict \emph{Low}, both pass T2 with comparable Judge scores, yet only Gemini~3~Flash answers the structured cue-grounding probes correctly.

\begin{examplebox}[title={Task 2: open-ended reasoning (both models predict Low Extraversion)}]
\begin{itemize}[leftmargin=1.4em, itemsep=2pt, topsep=2pt]
\item \emph{GPT-4o} (T2-Avg4 $=\!7.8$): ``The person speaks in a calm and measured tone, with limited outward energy or enthusiasm, suggesting a preference for introspection over sociability. This behavior aligns with a lower level of extraversion, as they do not exhibit assertiveness or high talkativeness.''
\item \emph{Gemini 3 Flash} (T2-Avg4 $=\!8.2$): ``The speaker exhibits a low-energy vocal tone and frequent gaze aversion, which are characteristic of introversion and internal cognitive processing. His limited use of expansive gestures and soft speech further support a lower level of extraversion.''
\end{itemize}
\textcolor{gray}{The two rationales are surface-level very similar. Both correctly arrive at \emph{Low Extraversion}, both cite generic cues (calm tone, low energy). The Judge gives them within $0.5$ points of each other.}
\end{examplebox}

\begin{examplebox}[title={Task 3: structured cue-grounding probe (same trait, divergent outcome)}]
\emph{Q.} During $4.9$\,s\,--\,$8.7$\,s, the person exhibits a specific behavioral pattern. Which Big Five personality dimension is most directly supported by this behavior?
\begin{itemize}[leftmargin=1.4em, itemsep=0pt, topsep=1pt, parsep=0pt]
\item[A.] \emph{Low Extraversion due to an internally-focused cognitive processing style} \hfill (\emph{correct})
\item[B.] High Agreeableness because the action shows a patient and calm demeanor.
\item[C.] High Conscientiousness due to the deliberate and careful thought process.
\item[D.] High Extraversion as the person is actively engaging with their thoughts.
\item[E.] Low Neuroticism because the behavior indicates a high level of comfort.
\item[F.] Low Openness since the behavior suggests a resistance to new information.
\end{itemize}
\textcolor{openc}{\textbf{GPT-4o picked C}} (incorrect). \quad \textcolor{humanc}{\textbf{Gemini 3 Flash picked A}} (correct).
\end{examplebox}

The verified observation at $[4.9, 8.7]$\,s reads ``the subject's gaze drifts down and to the left while the face remains neutral and speech continues,'' a textbook indicator of internally-directed cognitive processing and therefore of Low Extraversion. Both models predicted \emph{Low Extraversion} at the rating level, so the GT trait label was within reach for both. But only Gemini~3~Flash can localize \emph{which behavioral window} actually anchors the rating. GPT-4o's matching T1 + plausible T2 reasoning are not, on this sample, supported by retrievable cues, the very pattern the Prejudice Rate is designed to detect. The same divergence is visible across the remaining six MCQs of this video, which span Counterfactual / Mixed-Emotion / Spatial / Temporal-Spatial / Micro-expression / Temporal-Causal categories.

\section{Closed-vs-Open Frontier-Mean Task Gap}
\label{app:eco_gap}

Table~\ref{tab:gap_inversion} reports the frontier-mean (top-3 within each ecosystem) by task, referenced from \S\ref{sec:gap_inversion}. Closed leads on every task, but the gap is task-dependent: small on T1 and T2, several times larger on T3. Read together with the Prejudice Rate distribution (Figure~\ref{fig:pr_t1_scatter}), this confirms that the \gapinv{} is a field-wide phenomenon whose ecosystem-level component concentrates on cue retrieval rather than rating or verbal reasoning.

\begin{table}[t]
\caption{\textbf{Closed-vs-open frontier-mean task gap.} Top-3 within each ecosystem averaged by task. $\Delta$\,(abs.) is the absolute difference in percentage points (pp); $\Delta$\,(rel.) is the corresponding relative percent change.}
\label{tab:gap_inversion}
\centering
\small
\setlength{\tabcolsep}{9pt}
\renewcommand{\arraystretch}{1.2}
\scalebox{\tabgapscale}{%
\begin{tabular}{lccccc}
\toprule
\textbf{Metric} & \textbf{Top-3 Closed} & \textbf{Top-3 Open} & \textbf{$\Delta$\,(abs.)} & \textbf{$\Delta$\,(rel.)} & \textbf{Winner} \\
\midrule
T1 Accuracy (\%)        & \best{59.13} & 55.80 & $-3.33$\,pp  & $-5.6\%$  & \textcolor{closedc}{Closed (narrow)}  \\
T2-Avg4 (1--10)         & \best{6.63}  & 6.39  & $-0.24$      & $-3.6\%$  & \textcolor{closedc}{Closed (narrow)} \\
T3 Accuracy (\%)        & \best{67.84} & 49.77 & $-18.06$\,pp & \best{$-26.6\%$} & \textcolor{closedc}{Closed (large)} \\
\bottomrule
\end{tabular}%
}
\end{table}

\section{Generation-over-Time Effects}
\label{app:gen_effects}

\begin{figure}[t]
\centering
\includegraphics[width=0.92\linewidth]{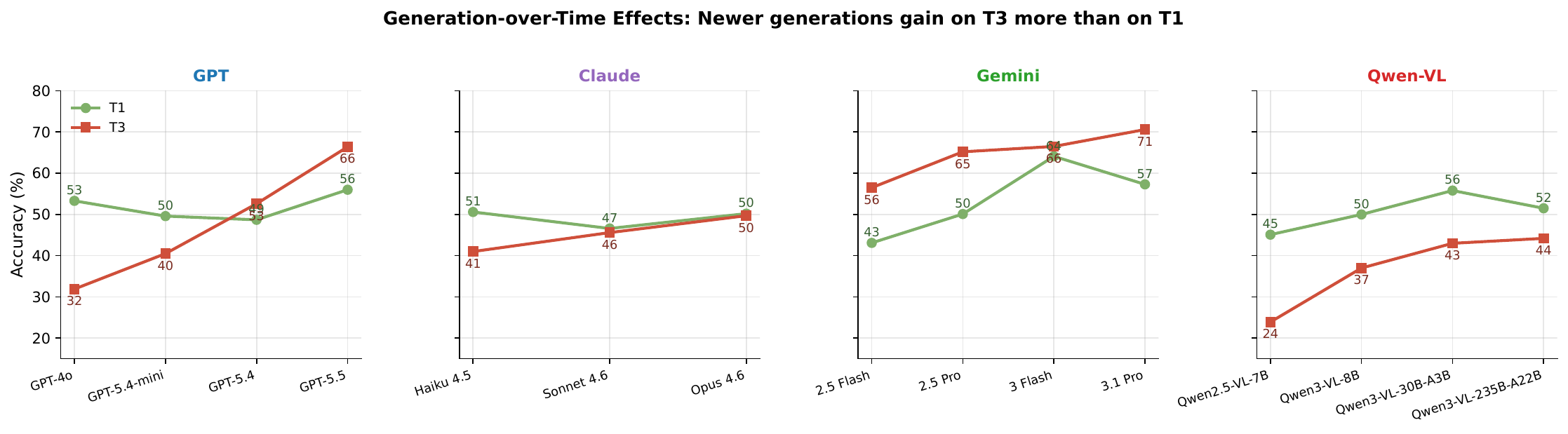}
\caption{\textbf{Generation-over-time per family.} T3 (cue grounding) improves much more steeply across each family's generations than T1 (rating), which saturates quickly.}
\label{fig:gen_effects}
\end{figure}

Within each closed-model family, every new generation improves T3 substantially while T1 saturates earlier (Figure~\ref{fig:gen_effects}). GPT-4o $\to$ GPT-5.5: T3 jumps from $31.9\%$ to $66.4\%$ ($+34.5$\,pp). Claude Haiku $\to$ Sonnet $\to$ Opus: T3 climbs from $41.0\%$ to $49.7\%$. Gemini 2.5 Pro $\to$ 3.1 Pro: T3 from $65.2\%$ to $70.6\%$. Within Gemini, the \emph{Flash} variants outperform \emph{Pro} on T1 (Gemini~3 Flash $64.1\%$ vs.\ 3.1 Pro $57.3\%$), while Pro variants outperform on T3 (3.1 Pro $70.6\%$ vs.\ 3 Flash $66.5\%$). Higher inference budget appears most useful for multi-step MCQ reasoning and less so for fast intuitive ratings. Open-source families (Qwen-VL 2.5\,$\to$\,3) show the same direction. Each generation roughly halves the gap to closed Top-3.

\section{Positional Bias}
\label{app:posbias}

\begin{figure}[t]
\centering
\includegraphics[width=0.7\linewidth]{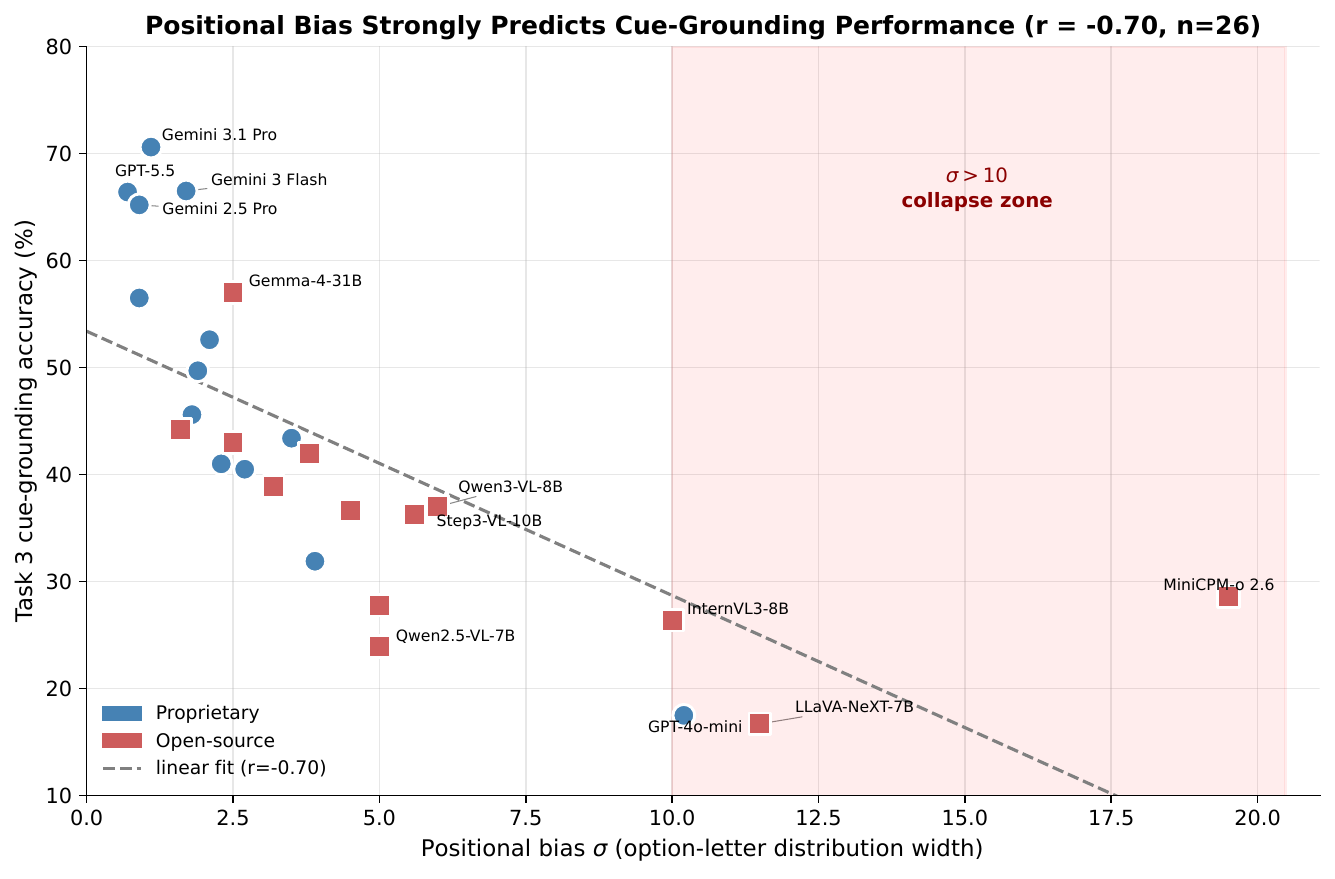}
\caption{\textbf{Positional bias $\sigma$ vs.\ Task~3 accuracy} across the 27 evaluated MLLMs. Models with $\sigma\!>\!10$ collapse to the bottom third on T3, regardless of family or scale.}
\label{fig:posbias}
\end{figure}

We measure positional bias as
\begin{equation}
\sigma(m) = \sqrt{\tfrac{1}{6}\sum_{\ell\in\{\texttt{A},\dots,\texttt{F}\}}\!\left(p_\ell(m) - \bar{p}\right)^2}\times 100, \quad \bar{p}=16.\overline{6}\%,
\label{eq:sigma}
\end{equation}
where $p_\ell(m)$ is the fraction of MCQs for which model $m$ selects option letter $\ell$. Across the 27 models, the top three with the lowest $\sigma$ are GPT-5.5 ($0.7$), Gemini~2.5 Flash ($0.9$), Gemini~2.5 Pro ($0.9$); the worst are MiniCPM-o~2.6 ($19.5$), LLaVA-NeXT ($11.5$), GPT-4o-mini ($10.2$), InternVL3-8B ($10.0$). \emph{Every model with $\sigma\!>\!10$ ranks in the bottom third on T3.} Figure~\ref{fig:posbias} plots $\sigma$ against T3 accuracy across the 27 models; the strong negative correlation ($r\!\approx\!-0.68$) makes positional-bias $\sigma$ a cheap early-warning signal of cue-retrieval collapse.

\section{Inter-Model Rank Correlation}
\label{app:rank_corr}

\begin{figure}[t]
\centering
\includegraphics[width=0.95\linewidth]{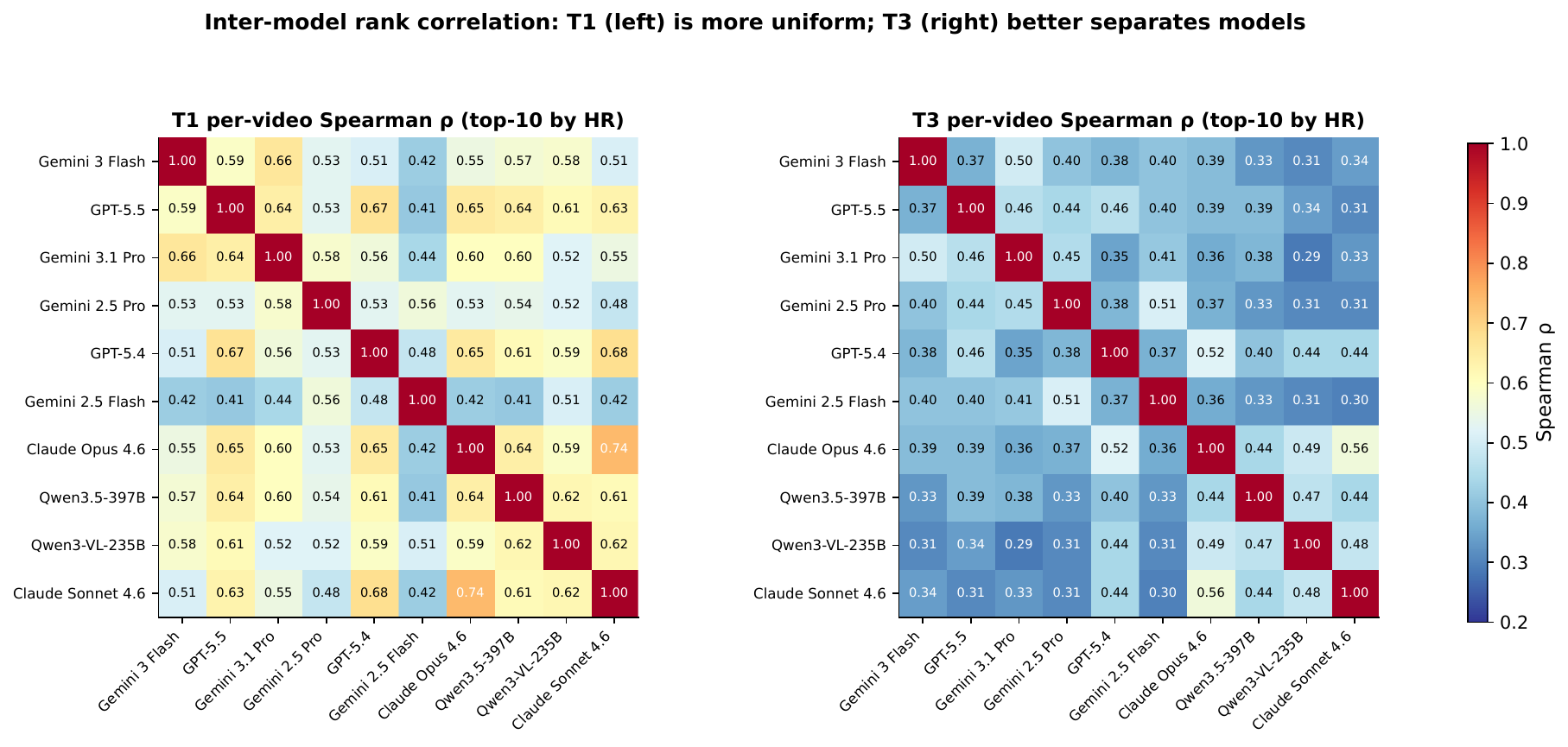}
\caption{\textbf{Pairwise per-video Spearman rank correlation among the Top-10 (by HR) MLLMs.} Left: T1 (rating); right: T3 (cue grounding). The lower-saturation T3 panel confirms that T3 carries less shared video-difficulty signal and more model-specific competence signal than T1.}
\label{fig:rank_corr}
\end{figure}

For each pair of models in the top-10 (by HR), we compute the Spearman rank correlation of per-video task scores. Mean off-diagonal $\rho$ for T1 is $0.56$ (range $0.41$--$0.74$); for T3 it is $0.39$ (range $0.29$--$0.56$). \emph{Top models agree more on which videos are intrinsically easy or hard to rate (T1) than on which questions they answer correctly (T3).} This confirms that T3 separates per-model competence more sharply than T1 (which carries a stronger shared video-difficulty signal), making T3 the cleaner discriminator and HR (which conditions on T3) the sharpest combined metric. Figure~\ref{fig:rank_corr} shows the full $10\!\times\!10$ correlation matrices for T1 and T3.

\section{Compute Resources}
\label{app:compute}

All open-source models are served on NVIDIA H200 GPUs via vLLM~\cite{kwon2023vllm}; all proprietary models are accessed through official APIs. Table~\ref{tab:compute} summarizes the estimated compute for each phase of the project.

\begin{table}[t]
\caption{\textbf{Estimated compute resources.}}
\label{tab:compute}
\centering
\small
\setlength{\tabcolsep}{6pt}
\renewcommand{\arraystretch}{1.15}
\begin{tabular}{lrr}
\toprule
\textbf{Phase} & \textbf{GPU (H200-hrs)} & \textbf{API cost (USD)} \\
\midrule
\multicolumn{3}{l}{\textit{Dataset construction}} \\
\addlinespace[2pt]
Annotation pipeline (Observer/Psychologist/Examiner/Aligner) & -- & $\sim$\$200 \\
Human annotation (24 annotators, 45K clues) & -- & -- \\
\addlinespace[4pt]
\multicolumn{3}{l}{\textit{Main benchmark evaluation}} \\
\addlinespace[2pt]
14 open-source models $\times$ 1,104 videos (unified prompt) & $\sim$60 & -- \\
13 proprietary models $\times$ 1,104 videos (API) & -- & $\sim$\$400 \\
AI-as-Judge (GPT-4o-mini, $\sim$149K calls) & -- & $\sim$\$120 \\
\addlinespace[4pt]
\multicolumn{3}{l}{\textit{Robustness and auxiliary experiments}} \\
\addlinespace[2pt]
Cross-judge robustness (Haiku + Flash-Lite, 200-video subset) & -- & $\sim$\$60 \\
Threshold sensitivity sweep (27 combos, offline recomputation) & $<$1 & -- \\
Preliminary / debugging / failed runs ($\sim$50\% overhead) & $\sim$30 & $\sim$\$100 \\
\midrule
\textbf{Total} & $\sim$\textbf{90} & $\sim$\textbf{\$880} \\
\bottomrule
\end{tabular}
\end{table}

\end{document}